\documentclass[twoside]{article}

\usepackage[accepted]{aistats2025}
\usepackage{alias}
\begin{document}

%

%

\twocolumn[

\aistatstitle{Bayesian Off-Policy Evaluation and Learning for Large Action Spaces}

\aistatsauthor{ Imad Aouali \And Victor-Emmanuel Brunel \And  David Rohde \And Anna Korba}

\aistatsaddress{ Criteo AI Lab \\ CREST, ENSAE, IP Paris \And  CREST, ENSAE, IP Paris \And Criteo AI Lab \And CREST, ENSAE, IP Paris} ]


\begin{abstract}
In interactive systems, actions are often correlated, presenting an opportunity for more sample-efficient off-policy evaluation (OPE) and learning (OPL) in large action spaces. We introduce a unified Bayesian framework to capture these correlations through structured and informative priors. In this framework, we propose \alg, a generic Bayesian approach for OPE and OPL, grounded in both algorithmic and theoretical foundations. Notably, \alg leverages action correlations without compromising computational efficiency. Moreover, inspired by online Bayesian bandits, we introduce Bayesian metrics that assess the average performance of algorithms across multiple problem instances, deviating from the conventional worst-case assessments. We analyze \alg in OPE and OPL, highlighting the benefits of leveraging action correlations. Empirical evidence showcases the strong performance of \alg.
\end{abstract}

\section{INTRODUCTION}\label{sec:introduction}
An off-policy contextual bandit \citep{dudik2011doubly} provides a practical framework for optimizing decision-making using logged data from an agent's interactions with an online environment \citep{bottou2013counterfactual}. The data, comprising context-action-reward tuples, is generated when the agent observes a context, takes an action, and receives a reward based on that action and context. This data allows two key tasks: off-policy evaluation (OPE) \citep{dudik2011doubly}, which estimates the expected reward of a new policy, and off-policy learning (OPL) \citep{swaminathan2015batch}, which learns an improved policy. Many interactive systems, like online advertising, can be modeled this way, where the context is user features, the action is product selection, and the reward is the click-through rate.

For small action spaces, existing methods based on the inverse propensity scoring (IPS) estimator \citep{horvitz1952generalization} are effective and come with theoretical guarantees. However, in large action spaces, these methods suffer from high bias and variance. The bias is due to the logging policy's\footnote{The logging policy is the policy of the agent used to collect logged data.} deficient support \citep{sachdeva2020off}, while the variance is due to high importance weight values \citep{swaminathan2017off}. To address this, \citet{saito2022off} introduced MIPS, an IPS variant for OPE in large action spaces. MIPS relies on informative embeddings (and their distribution) that fully capture the causal effects of actions on rewards. Numerous recent variants of MIPS have been proposed \citep{peng2023offline,sachdeva2023off,cief2024learning,taufiq2024marginal,saito2023off} to relax the original assumptions made by MIPS. All these studies are variants of IPS, designed to capture some underlying structure of the problem under different/milder assumptions than MIPS.

While much of the existing work focuses on IPS, an alternative is the direct method (DM) \citep{jeunen2021pessimistic}, which directly learns a reward model. DMs typically introduce less variance than IPS but can be more prone to bias if the model is poorly suited to the data. DMs are widely used in practice, particularly in recommender systems \citep{sakhi2020blob,jeunen2021pessimistic,aouali2022probabilistic}, but their theory is less explored. This work seeks to analyze DMs from a Bayesian perspective and improve their efficiency in OPE and OPL by incorporating informative priors, especially in large action spaces. Similar to how MIPS \citep{saito2022off} extended IPS with auxiliary information, we aim to enhance DMs by integrating such information. However, unlike MIPS’s focus on OPE, we offer theoretical and empirical insights for both OPE and OPL, relying on different assumptions than MIPS and its variants.

\textbf{Contributions.} We introduce a Bayesian method, called structured Direct Method (\alg), which leverages informative priors to share reward information across actions. When one action is observed, \alg updates its knowledge about similar actions, improving statistical efficiency without compromising computational scalability, making it suitable for large action spaces: a key limitation of traditional DMs. To evaluate \alg, we introduce Bayesian metrics for OPE and OPL that measure average performance across different instances, diverging from the frequentist worst-case focus. These metrics capture the benefits of informative priors, enhancing our understanding of DMs, especially given the relatively underdeveloped statistical analysis of DMs in OPE and OPL. Finally, we evaluate \alg using both synthetic and real data.

\textbf{Related work.} Bayesian models for OPL have been studied by \citet{lazaric2010bayesian,hong2023multi}, but these works focused on multi-task learning. Moreover, the models they use differ from ours. \citet{hong2023multi} also introduced a Bayesian metric for OPL that evaluates performance in a single environment drawn from the prior given a fixed logged dataset. In contrast, our metric assesses the average performance across all environments sampled from the prior and all logged datasets, which is more typical in Bayesian analyses (e.g., Bayesian regret in \citet{russo14learning}). Furthermore, we cover OPE, expanding beyond the OPL-focused scope of \citet{lazaric2010bayesian,hong2023multi}. Finally, our theoretical results are more general and rely on much weaker assumptions. For more related works, refer to \cref{sec:extended_related_work}.
\section{BACKGROUND}
\label{sec:setting}
Let $[n] = \{1, \dots ,n\}$ for any positive integer $n$. Random variables are denoted with capital letters, and their realizations with the respective lowercase letters, except for Greek letters. Let $X, Y$ be two random variables, with a slight abuse of notation, the distribution (or density) of $X \mid Y=y$ evaluated at $x$ is denoted by $p(x \mid y)$. Let $a_1, \ldots, a_n$ be $n$ vectors of $\real^d$, $a = (a_i)_{i \in [n]} \in \real^{nd}$ is their $nd$-dimensional concatenation, $\otimes$ denotes the Kronecker product, and $\mathcal{O}$ the big-O notation. For any vector $a \in \real^d$ and any positive-definite matrix $\Sigma \in \real^{d\times d}$, we define $\normw{a}{\Sigma} = \sqrt{a^\top \Sigma a}$. Finally, $\lambda_1(A)$ and $\lambda_d(A)$ denote the maximum and minimum eigenvalue of matrix $A$, respectively.

Agents are represented by stochastic policies $\pi \in \Pi$, where $\Pi$ is the set of policies. Precisely, given a context $x \in \real^{d}$, $\pi(\cdot \mid  x)$ is a probability distribution over a finite \emph{action set} $\cA = [K]$, where $K$ is the number of actions. Agents interact with a \emph{contextual bandit} environment over $n$ rounds. In round $i \in [n]$, the agent observes a \emph{context} $X_i \sim \nu$, where $\nu$ is a distribution whose support $\mathcal{X}$ is a compact subset of $\mathbb{R}^{d}$. Then the agent takes an \emph{action} $A_i \sim \pi_0(\cdot \mid X_i)$ from the action set $\cA$, where $\pi_0$ is the \emph{logging policy}. Finally, assuming a parametric model, the agent receives a \emph{stochastic reward} $R_i \sim p(\cdot \mid  X_i; \theta_{*, A_i})$. Here $p(\cdot \mid  x; \theta_{*, a})$ is the \emph{reward distribution} for action $a$ in context $x$, where $\theta_{*, a} \in \real^d$ is an \emph{unknown $d$-dimensional parameter} of action $a$. Let $\theta_{*} = (\theta_{*, a})_{a \in \cA} \in \real^{dK}$ be the concatenation of action parameters. Then, $r(x, a; \theta_*) = \E{R \sim p(\cdot \mid  x; \theta_{*, a})}{R}$ is the \emph{reward function} that outputs the expected reward of action $a$ in context $x$. Finally, the goal is to find a policy $\pi \in \Pi$ that maximizes the value function $V(\pi; \theta_*) =\mathbb{E}_{X \sim \nu}\E{A \sim \pi(\cdot | x)}{r(X, A; \theta_*)}$.

Let $S = (X_i, A_i, R_i)_{i \in [n]}$  be a set of random variables drawn i.i.d. as $X_i \sim \nu$, $A_i \sim \pi_0(\cdot \mid X_i)$ and $R_i \sim p(\cdot \mid  X_i; \theta_{*, A_i})$; that we will refer to as the sample set. The goal of off-policy evaluation (OPE) and learning (OPL) is to build an estimator $\hat{V}(\pi, S)$ of $V(\pi; \theta_*)$ and then use $\hat{V}(\pi, S)$ to find a policy $\hat{\pi} \in \Pi$ that maximizes $V(\cdot;\theta_*)$. In OPE, it is common to use \emph{inverse propensity scoring (IPS)} \citep{horvitz1952generalization,dudik2012sample}, which leverages importance sampling to estimate the value $V(\pi;\theta_*)$ as $\hat{V}_{\textsc{ips}}(\pi, S) = \frac{1}{n} \sum_{i \in [n]} \frac{\pi(A_i \mid  X_i)}{\pi_0(A_i \mid  X_i)} R_i$
where for any $(x, a) \in \cX \times \cA\,, \frac{\pi(a\mid  x)}{\pi_0(a\mid x)}$ are the \emph{importance weights}. 
In practice, IPS can suffer high variance, especially when the action space is large \citep{swaminathan2017off, saito2022off}. Moreover, while IPS is unbiased under the assumption that the logging policy has full support, it can induce a high bias when such an assumption is violated \citep{sachdeva2020off}, which is again likely when the action space is large. 
IPS, also, assumes access to the logging policy $\pi_0$. An alternative approach to IPS is to use a \emph{direct method (DM)} \citep{jeunen2021pessimistic}, that relies on a reward model $\hat{r}$ to estimate the value $V(\pi; \theta_*)$ 
 as
\begin{align}\label{eq:dm_policy_value}
    \hat{V}_{\textsc{dm}}(\pi, S) &= \frac{1}{n} \sum_{i \in [n]} \sum_{a \in \cA} \pi(a \mid  X_i) \hat{r}(X_i, a)\,,
\end{align}
where $\hat{r}(x, a)$ is an estimation of $r(x, a; \theta_*)$. DM estimators may exhibit modeling bias, but they generally have lower variance than IPS \citep{saito2022off}. Another advantage of DM is its practical utility without assuming access to the logging policy $\pi_0$ \citep{jeunen2021pessimistic,aouali2022probabilistic,hong2023multi}. Also, DMs can be incorporated into a Bayesian framework, where informative priors can be used to enhance statistical efficiency. This allows for the development of scalable methods suitable for large action spaces, as shown in our work. 



\section{STRUCTURED DM}\label{sec:model}
\subsection{Structured Priors}
\textbf{Pitfalls of non-structured priors.} Before presenting \alg, we first describe the pitfalls of using the following widely used standard prior,
\begin{align}\label{eq:basic_model}
  \theta_a &\sim \cN(\mu_{a}, \Sigma_{a})\,, & \forall a \in \cA\,,\\
    R \mid \theta, X, A &\sim \cN(\phi(X)^\top \theta_A , \sigma^2)\,,\nonumber
\end{align}
where \( \phi(x) \) provides a \( d \)-dimensional representation of the context \( x \in \mathcal{X} \), and \( \mathcal{N}(\mu_{a}, \Sigma_{a}) \) represents the prior density of \( \theta_a \), with \( \sigma^2 \) being the reward noise. Under this prior, each action \( a \) has an associated parameter \( \theta_a \). Given the prior in \eqref{eq:basic_model}, the posterior distribution of an action parameter follows a multivariate Gaussian: \( \theta_a \mid S \sim \mathcal{N}(\hat{\mu}_a, \hat{\Sigma}_a) \), where \( \hat{\Sigma}_{a}^{-1} = \Sigma_{a}^{-1} + G_{a} \) and \( \hat{\Sigma}_{a}^{-1} \hat{\mu}_{a} = \Sigma_{a}^{-1} \mu_a + B_a \). Here, \( G_{a} = \sigma^{-2} \sum_{i \in [n]} \mathbb{I}_{\{A_i=a\}} \phi(X_i) \phi(X_i)^\top \) and \( B_{a} = \sigma^{-2} \sum_{i \in [n]} \mathbb{I}_{\{A_i=a\}} R_i \phi(X_i) \). Note that \( G_a \) and \( B_a \) only use the subset of samples \( S \) where action \( a \) was observed, meaning data from other actions \( b \neq a \) do not contribute to the posterior inference for action \( a \). This results in statistical inefficiency, especially if the sample set \( S \) doesn't cover all actions. In particular, the posterior for an unseen action \( a \), \( \theta_a \mid S \), would simply revert to the prior \( \mathcal{N}(\mu_a, \Sigma_a) \), since we would have \( G_a = 0_{d \times d} \) and \( B_a = 0_d \) in such case.

\textbf{Structured priors.} To address the above issue, we assume that action rewards correlate and embed this knowledge into the prior. While one could model these correlations by considering the joint posterior distribution of \( (\theta_a)_{a \in \mathcal{A}} \mid S \), this becomes computationally burdensome when the number of actions \( K \) is large. Instead, we introduce an \emph{unknown $d^\prime$-dimensional latent parameter} \( \psi \in \mathbb{R}^{d^\prime} \), sampled from a \emph{latent prior} \( q(\cdot) \), such as \( \psi \sim q(\cdot) \). The correlations between actions naturally arise because each action parameter \( \theta_a \) is derived from the same latent parameter \( \psi \).

Specifically, the action parameters \( \theta_a \) are conditionally independent given \( \psi \) and are sampled from a \emph{conditional prior} \( p_a \) as \( \theta_a \mid \psi \sim p_a(\cdot ; f_a(\psi)) \) for all \( a \in \mathcal{A} \). Here, \( p_a \) is parameterized by \( f_a(\psi) \), where \( f_a: \mathbb{R}^{d^\prime} \rightarrow \mathbb{R}^d \) is a known prior function that encodes the hierarchical relationship between action parameters \( \theta_a \) and the latent parameter \( \psi \). This structure allows for sparsity, meaning that \( \theta_a \) may depend only on a subset of \( \psi \)'s coordinates. Moreover, \( p_a \) accounts for model uncertainty, allowing for cases where \( \theta_a \) is not a deterministic function of \( \psi \), i.e., \( \theta_a \neq f_a(\psi) \).

\begin{figure}[H]
\begin{center}
\includegraphics[width=\linewidth]{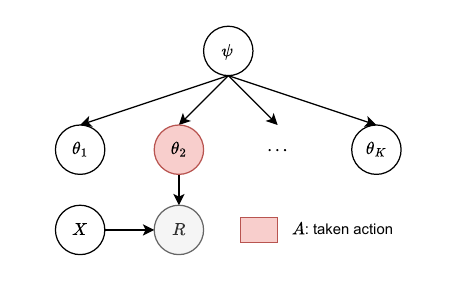}
\caption{Graph representation of the structured prior.}
\label{fig:sdm_graph}
\end{center}
\end{figure}

The reward distribution for action \( a \) in context \( x \) is given by \( p(\cdot \mid x; \theta_a) \), which depends only on \( x \) and \( \theta_a \). To summarize, the structured prior is defined below, and its graphical representation is given in \cref{fig:sdm_graph}.
\begin{align}\label{eq:model}
  \psi &\sim q(\cdot)\,, \\
  \theta_a \mid  \psi &\sim p_{a}(\cdot ; f_a(\psi))\,, &  \forall a \in \mathcal{A}\,, \nonumber \\
  R \mid \psi, \theta, X, A &\sim p(\cdot \mid  X; \theta_A)\,. \nonumber
\end{align}
This prior is flexible as assuming the existence of a latent parameter \( \psi \) is a mild assumption (\cref{subsec:assumptions}). To derive the posterior under this prior, we assume that: \emph{(i)} \( (X, A) \) is independent of \( \psi \), and given \( \psi \), \( (X, A) \) is independent of \( \theta \); and \emph{(ii)} given \( \psi \), the parameters \( \theta_a \) for all \( a \in \mathcal{A} \) are independent.

Now, we discuss how to perform OPE and OPL under this general structured prior in \eqref{eq:model}, before applying it to linear-Gaussian distributions in \cref{sec:application}.

\subsection{Off-Policy Evaluation and Learning}\label{subsec:ope}
\textbf{Off-policy evaluation.} OPE aims at estimating the value function \(V(\pi; \theta_*)\) using the sample set \(S\). In DMs, the estimator \(\hat{V}_{\textsc{dm}}\) in \eqref{eq:dm_policy_value} requires access to the learned reward \(\hat{r}(x, a) \approx r(x, a; \theta_*)\). In our Bayesian setting, this requires access to the action posterior \(\theta_{a} \mid S\) under model \eqref{eq:model} since the reward is then estimated as \(\hat{r}(x, a) = \condE{r(x, a; \theta)}{S}\) for any \((x, a) \in \cX \times \cA\), and this estimate is plugged into \(\hat{V}_{\textsc{dm}}\) in \eqref{eq:dm_policy_value} to estimate \(V(\pi; \theta_*)\). Thus, we need to derive the posterior density of the action parameter \(\theta_a\), \(p(\theta_a \mid S)\), under the structured prior in \eqref{eq:model}, which reads
\begin{align}\label{eq:action_posterior}
  p(\theta_a \mid S)= \int_{\psi} p(\theta_a \mid \psi, S) p(\psi \mid S)
 \dif \psi\,,
\end{align}
where $\psi \mid S$ is the latent posterior and $\theta_a \mid \psi, S$ is the \emph{conditional} action posterior. To compute $p(\theta_a \mid S)$, we first compute $p(\theta_a \mid \psi, S)$ and $ p(\psi \mid S)$ and then integrate out $\psi$ following \eqref{eq:action_posterior}. First,
\begin{align}\label{eq:p_derviation}
 p(\theta_a \mid \psi, S) \propto \LL_{a}(\theta_a) p_a (\theta_a ; f_a(\psi))\,, 
\end{align}
with $\LL_{a}(\theta_a) = \prod_{(X, A, R) \in S_a} p(R| X; \theta_a)$ is the likelihood of observations of action $a$ ($S_a = (X_i, A_i, R_i)_{i \in [n], A_i=a}$ is the subset of $S$ where $A_i=a$). Similarly,
\begin{align}\label{eq:q_derviation}
\hspace{-0.1cm} p(\psi \mid S) & \propto \prod_{b \in \cA} \int_{\theta_b} \LL_{b}(\theta_b) p_b\left(\theta_b ; f_b(\psi) \right) \dif \theta_b \, q(\psi)\,, 
\end{align}
This allows us to further develop \eqref{eq:action_posterior} as
\begin{align}\label{eq:action_posterior_summary}
  p(\theta_a \mid S)  \propto &  \int_{\psi}\LL_{a}(\theta_a) p_a (\theta_a ; f_a(\psi)) \\ & \prod_{b \in \cA} \int_{\theta_{b}} \LL_{b}(\theta_{b}) p_{b}\left(\theta_{b} ; f_{b}(\psi) \right) \dif \theta_{b}  \, q(\psi) \dif \psi\,.\nonumber
\end{align}
All the quantities inside the integrals in \eqref{eq:action_posterior_summary} are given (the parameters of $p_a$ and $q$) or tractable (the terms in $\LL_{a}$). Thus, if these integrals can be computed, then the posterior can be fully characterized in closed-form which we will do in \cref{sec:application} in the fully linear case. 
Otherwise, the posterior should be approximated, e.g. through approximate sampling methods, see \cref{subsec:steps_lin_hierarchy_nonlin_rewards} for more details.


\textbf{Greedy off-policy learning.} OPL aims at finding a policy $\pi$ that maximizes the value $V(\pi ;\theta_*)$. We act greedy with respect to our estimator $\hat{V}_{\textsc{dm}}(\cdot, S)$ and define the learned policy as the one maximizing it: $\hat{\pi}_{\textsc{g}} = \argmax_{\pi \in \Pi} \hat{V}_{\textsc{dm}}(\pi, S)$. If the set of policies $\Pi$ contains deterministic policies, then 
\begin{align}\label{eq:learning_procedure_2}
     \hat{\pi}_{\textsc{g}}(a \mid x) = \mathds{1}{\{a = \argmax_{b \in \cA} \hat{r}(x, b)\}}\,.
\end{align}
In particular, we do not adopt the common pessimism approach \citep{jin2021pessimism}. In pessimism, one constructs confidence intervals of the reward estimate $\hat{r}(x, a)$ of the form $|r(x, a; \theta) - \hat{r}(x, a)| \leq u(x, a)$, and then defines the learned policy as $\hat{\pi}_{\textsc{p}}(a \mid x) = \mathds{1}{\{a = \argmax_{b \in \cA} \hat{r}(x, b) - u(x, b)\}}$. If $u(x, a)$ does not depend on $x$ and $a$, then pessimistic and greedy policies are the same. The advantage of one over another depends on the evaluation metric. Our metric is the Bayesian suboptimality (BSO), defined in \cref{sec:analysis}. It assesses the average performance of algorithms across multiple problems rather than focusing on the worst-case. The Greedy policy is more suitable for BSO optimization than pessimism (demontrated theoretically in \cref{sec:analysis} and empirically in \cref{app:greedy_pessimism_experiments}).

\section{LINEAR-GAUSSIAN CASE}\label{sec:application}
In this section, we use linear functions $f_a$ combined with Gaussian distributions for the structured prior~\eqref{eq:model}. Precisely, we assume that the latent prior $q(\cdot) = \cN(\cdot; \mu, \Sigma)$ is Gaussian with mean $\mu \in \real^{d^\prime}$ and covariance $\Sigma \in \real^{d^\prime \times d^\prime}$. Moreover, let $\W_a \in \real^{d \times d^\prime}$ be the \emph{mixing matrix} for action $a$, we define $f_a(v) = \W_a v$ for any $v \in \real^{d^\prime}$. We define the conditional prior $p_{a}(\cdot ; f_a(\psi)) = \cN(\cdot; \W_a \psi, \Sigma_a)$ is Gaussian with mean $f_a(\psi) = \W_a \psi \in \real^d$ and covariance $\Sigma_{a} \in \real^{d \times d}$. The reward distribution $p(\cdot \mid x; \theta_a)$ is also linear-Gaussian as $\cN(\cdot; \phi(x)^\top \theta_a, \sigma^2)$, where $\phi(\cdot)$ outputs
a $d$-dimensional representation of $x$ and $\sigma>0$ is the observation noise. The whole prior is 
\begin{align}\label{eq:contextual_gaussian_model}
    \psi & \sim \cN(\mu, \Sigma)\,, \\ 
    \theta_a \mid   \psi& \sim \cN\Big( \W_a \psi , \,  \Sigma_{a}\Big)\,, & \forall a \in \cA\,,\nonumber\\ 
    R \mid  \psi, \theta, X, A & \sim \cN(\phi(X)^\top \theta_A ,  \sigma^2)\,. \nonumber
\end{align}
This is an important model, in both practice and theory since linear models often lead to closed-form solutions that are both computationally tractable and allow theoretical analysis. To highlight the generality of \eqref{eq:contextual_gaussian_model}, we provide some problems where it can be used.

\subsection{Applications}
\textbf{Mixed-effect modeling.} \eqref{eq:contextual_gaussian_model} allows modeling that action parameters depend on a linear mixture of effect parameters. Precisely, let $J$ be the number of effects and assume that $d^\prime = dJ$ so that the latent parameter $\psi$ is the concatenation of $J$, $d$-dimensional effect parameters, $\psi_j \in \real^d$, such as $ \psi = (\psi_j)_{j \in [J]} \in \real^{dJ}$. Moreover, assume that for any $a \in \cA\,,$ $\W_a=w_a^\top \otimes I_d \in \real^{d \times dJ}$ where $w_a = (w_{a, j})_{j \in [J]} \in \mathbb{R}^{J}$ are the \emph{mixing weights} of action $a$. Then, $\W_a \psi = \sum_{j \in [J]} w_{a, j} \psi_j$ for any $a \in \cA$. Sparsity, i.e., when an action $a$ only depends on a subset of effects, is captured through the mixing weights $w_a$: $w_{a, j}=0$ when action $a$ is independent of the $j$-th effect parameter $\psi_j$ and $w_{a, j}\ne 0$ otherwise. Also, the level of dependence between action $a$ and effect $j$ is quantified by the absolute value of $w_{a, j}$. This mixed-effect model can be used in numerous applications. In movie recommendation, $\theta_a$ is the parameter of movie $a$, $\psi_j$ is the parameter of theme $j$ (adventure, romance, etc.), and $w_{a, j}$ quantifies the relevance of theme $j$ to movie $a$. Similarly, in clinical trials, a drug is a combination of multiple ingredients, each with a specific dosage. Then $\theta_a$ is the parameter of drug $a$, $\psi_j$ is the parameter of ingredient $j$, and $w_{a, j}$ is the dosage of ingredient $j$ in drug $a$.

\textbf{Low-rank modeling.} \eqref{eq:contextual_gaussian_model} can also model the case where the dimension of the latent parameter $\psi$ is much smaller than that of the action parameters $\theta_a$, i.e., when $d^\prime \ll d$. Again, this is captured through the mixing matrices $\W_a$, when $\W_a$ is low-rank.

\subsection{Closed-Form Solutions for \alg}\label{subsec:steps_lin_hierarchy}
The conditional action posterior  is known in closed-form as 
$\theta_a \mid \psi, S \sim \cN(\tilde{\mu}_{a}, \tilde{\Sigma}_{a})$, with
\begin{align}\label{eq:conditional_posterior}
   &\tilde{\Sigma}_{a}^{-1}  = \Sigma_{a}^{-1} + G_{a}\,, &   \tilde{\Sigma}_{a}^{-1} \tilde{\mu}_{a} =  \Sigma_{a}^{-1} \W_a \psi + B_{a}\,,
\end{align}
where $G_{a} = \sigma^{-2} \sum_{i \in [n]} \mathds{1}{\{A_i=a\}} \phi(X_i) \phi(X_i)^\top$ and $B_{a} = \sigma^{-2} \sum_{i \in [n]} \mathds{1}{\{A_i=a\}} R_i \phi(X_i)$. This posterior has the standard form except that the prior mean $\W_a \psi$ now depends on the latent parameter $\psi$. Similarly, the effect posterior writes $\psi \mid S \sim \cN(\bar{\mu}, \bar{\Sigma})$, where
\begin{align}\label{eq:effect_posterior}
   &\bar{\Sigma}^{-1} =   \Sigma^{-1}   + \sum_{a \in \cA} \W_a^{\top} ( \Sigma_a^{-1} - \Sigma_a^{-1} \tilde{\Sigma}_{a} \Sigma_a^{-1} )\W_a\,, \nonumber\\ 
   &\bar{\Sigma}^{-1}\bar{\mu} = \Sigma^{-1} \mu +\sum_{a \in \cA}   \W_a^\top \Sigma_a^{-1} \tilde{\Sigma}_{a}  B_a.
\end{align}
The latent posterior precision \( \bar{\Sigma}^{-1} \) is the sum of the latent prior precision \( \Sigma^{-1} \) and the learned action precisions \( \Sigma_a^{-1} - \Sigma_a^{-1} \tilde{\Sigma}_a \Sigma_a^{-1} \), weighted by \( \W_a^\top \W_a \). The contribution of each action's learned precision to the latent precision is proportional to \( \W_a^\top \W_a \). This intuition similarly applies to interpreting \( \bar{\mu} \). Finally, from \eqref{eq:action_posterior_summary}, the action posterior is \( \theta_a \mid S \sim \mathcal{N}(\hat{\mu}_a, \hat{\Sigma}_a) \), where
\begin{align}\label{eq:action_posterior_gaussian}
   &\hat{\Sigma}_a = \tilde{\Sigma}_{a}  +\tilde{\Sigma}_{a} \Sigma_{a}^{-1}  \W_a \bar{\Sigma} \W_a^\top \Sigma_{a}^{-1} \tilde{\Sigma}_{a}\,,\nonumber\\
   &\hat{\mu}_a = \tilde{\Sigma}_a \big( \Sigma_{a}^{-1} \W_a \bar{\mu} + B_{a} \big)  \,.
\end{align}
Finally, from \eqref{eq:contextual_gaussian_model}, the reward function is \( r(x, a; \theta) = \phi(x)^\top \theta \). Thus, the estimated reward is \( \hat{r}(x, a) = \mathbb{E}[r(x, a; \theta) \mid S] = \phi(x)^\top \hat{\mu}_a \) for any \( (x, a) \in \mathcal{X} \times \mathcal{A} \), which can then be plugged in \eqref{eq:learning_procedure_2} for decision-making, leading to $ \hat{\pi}_{\textsc{g}}(a \mid x) = \mathds{1}{\{a = \argmax_{b \in \cA} \phi(x)^\top \hat{\mu}_b\}}$.

To see why this is more beneficial than the standard prior \eqref{eq:basic_model}, notice that the mean and covariance of the posterior of action $a$, $\hat{\mu}_a$ and $\hat{\Sigma}_a$, are now computed using the mean and covariance of the latent posterior, $\bar \mu $ and $\bar \Sigma$. But $\bar \mu $ and $\bar \Sigma$ are learned using the interactions with all the actions in $S$. Thus $\hat{\mu}_a$ and $\hat{\Sigma}_a$ are also learned using the interactions with all the actions in $S$, in contrast with the standard prior \eqref{eq:basic_model} where they were learned using only the interaction with action $a$. The additional computational cost of considering the structured prior in \eqref{eq:contextual_gaussian_model} is small. The computational and space complexities are $\mathcal{O}(K((d^2 + {d^\prime}^2)(d + d^\prime)))$ and $\mathcal{O}(K d^2)$. For example, when $d^\prime = \mathcal{O}(d)$, these complexities become $\mathcal{O}(Kd^3)$ and $\mathcal{O}(Kd^2)$, respectively. This is exactly the cost of the standard prior in \eqref{eq:basic_model}. In contrast, this strictly improves the computational efficiency of jointly modeling the action parameters, where the complexities are $\mathcal{O}(K^3d^3)$ and $\mathcal{O}(K^2d^2)$ since the joint posterior of $(\theta_a)_{a \in \cA} \mid S$ requires converting and storing a $dK \times dK$ covariance matrix.



\begin{remark}
\alg with this linear Gaussian hierarchy can be used even with data generated from non-linear rewards, and we empirically investigate its robustness to misspecification. We found that this model performs well even if the true rewards are not generated from a linear-Gaussian distribution. An extension of \alg to binary rewards is provided in \cref{subsec:steps_lin_hierarchy_nonlin_rewards}.\end{remark}
\section{ANALYSIS}\label{sec:analysis}
While we address both OPL and OPE for our theoretical results, we place more emphasis on OPL and present it first for clarity. Although it might seem more natural to start with OPE, we found that this structure makes our contributions easier to explain. All proofs are provided in \cref{sec:ope_proofs}.

\subsection{Bayesian OPE and OPL Metrics}\label{subsec:bayes_suboptimality}
We introduce a Bayesian OPL metric, called Bayesian suboptimality. A similar OPE metric, Bayesian Mean Squared Error (BMSE), is presented afterwards.

\textbf{Bayesian Suboptimality.} In OPL, the performance of a learned policy \( \hat{\pi} \) is evaluated using suboptimality (SO): \( \textsc{so}(\hat{\pi}; \theta_*) = V(\pi_*; \theta_*) - V(\hat{\pi}; \theta_*) \), where \( \pi_* = \argmax_{\pi \in \Pi} V(\pi; \theta_*) \) is the optimal policy. This metric is well-suited when the environment is governed by a unique, fixed ground truth \( \theta_* \). It applies to any policy \( \hat{\pi} \), whether learned through frequentist approaches (e.g., MLE) or Bayesian ones (e.g., ours). However, when the environment is modeled as a random variable \( \theta_* \), SO becomes less appropriate. Thus, drawing on recent developments in Bayesian analysis for online bandits through Bayes regret \citep{russo14learning}, we introduce a new metric for offline settings, termed \emph{Bayes suboptimality}, defined as:
\begin{align}\label{eq:bayes_suboptimality} 
  \textsc{Bso}(\hat{\pi})=   \mathbb{E}[V(\pi_*; \theta_*) - V(\hat{\pi}; \theta_*)]\,,
\end{align}
where the expectation is taken over all random variables: the sample set \(S\) and \(\theta_*\), which is treated as a random variable sampled from the prior. The BSO can be computed in two ways. One method involves taking the expectation under the prior \(\theta_*\), followed by taking an expectation under data generated from a fixed environment \(\theta_*\) as \(S \mid \theta_*\). The other method involves taking an expectation under the data \(S\), followed by taking an expectation under the posterior \(\theta_* \mid S\). The BSO is a reasonable metric for assessing the average performance of algorithms across multiple environments, due to the expectation over \(\theta_*\). It is also known that Bayes regret captures the benefits of using informative priors \citep{aoualimixed}, and this is similarly achieved by the BSO (\cref{subsec:frequentist_vs_bayes}).

\citet{hong2023multi} also examined a concept of Bayesian suboptimality  defined as $V(\pi_*;\theta_*) - V(\hat{\pi};\theta_*) \mid S$, where the sample set $S$ is fixed and randomness only comes from $\theta_*$ sampled from the posterior. In contrast, our definition aligns with traditional online bandit settings (Bayesian regret \citep{russo14learning}) by averaging the suboptimality across both the sample set $S$ and parameter $\theta_*$ sampled from the prior. This leads to a different metric that captures the average performance under various data realizations and parameter draws. One major difference is that our metric favors the greedy policy (\cref{subsec:main_result}), while they use pessimism \citep[Sections 4 and 5]{hong2023multi}.

\textbf{Bayesian mean squared error (BMSE).} In OPE, we assess the quality of estimator $\hat{r}$ using the mean squarred error (MSE). Then, similarly to the BSO, we can define the BMSE of the estimator $\hat{r}$ for a fixed action $a$ and context $x$ as 
\begin{align}
    {\textsc{Bmse}}(\hat{r}(x, a)) 
    = \mathbb{E}\big[\big( \hat{r}(x, a) - r(x, a; \theta_*) \big)^2\big]\,.
\end{align}
The BMSE diverges from the MSE in that the expectation in BMSE is calculated over the sample set $S$ and across the true parameter $\theta_*$ sampled from the prior.


\subsection{Theoretical Results}\label{subsec:main_result}
Our theory relies on a well-specified prior assumption, where action parameters $\theta_{*, a}$ and true rewards are assumed to be drawn from the structured prior \eqref{eq:contextual_gaussian_model}. This yields our bound on the BSO of \alg.

\begin{theorem}[Covariance-Dependent Bound]\label{thm:main_thm_1}
Let $\pi_*(x)$ be the optimal action for context $x$. Then the BSO of \emph{\alg} under the structured prior \eqref{eq:contextual_gaussian_model} satisfies
\begin{align}
    \textsc{Bso}(\hat{\pi}_{\textsc{g}}) \leq 2\sqrt{d} \, \mathbb{E}\big[ \|\phi(X)\|_{\hat{\Sigma}_{\pi_*(X)}}\big]\,.
\end{align}
\end{theorem}
Scaling of the bound in \cref{thm:main_thm_1} aligns with existing frequentist results \citep[Theorem 4.4]{jin2021pessimism}. The main differences lie in the constants, which in our case reflect the benefits of using informative priors (\cref{subsec:frequentist_vs_bayes}), and the fact that this rate is achieved using greedy policies \eqref{eq:learning_procedure_2}. This contrasts with the frequentist setting where pessimism is used \citep{jin2021pessimism} and known to be optimal \citep[Theorem 4.7]{jin2021pessimism}. In fact, greedy policies are optimal when using BSO as a performance metric. Specifically, \(\textsc{Bso}(\hat{\pi}_{\textsc{g}}) \leq \textsc{Bso}(\pi)\) for any policy \(\pi\), including pessimistic ones \citep{jin2021pessimism}. Therefore, in the Bayesian setting and when using BSO as a performance metric, greedy policies should always be preferred to pessimistic ones. This fundamental difference is proven in \cref{subsec:greedy_pessimism} and it is of independent interest beyond this work. \cref{thm:main_thm_1} suggests that the BSO primarily depends on the posterior covariance of action $\pi_*(X)$ in the direction of the context $\phi(X)$. That is, when the uncertainty in the posterior distribution of the optimal action $\pi_*(X)$ is low on average across different contexts $X$, problem instances $\theta_*$, and sample sets $S$, then the BSO bound is correspondingly small. In particular, the tightness of the bound depends on the degree to which the sample set covers the optimal actions on average.

Even without such an assumption, \cref{thm:main_thm_1} still highlights the advantages of using \alg over the non-structured prior \eqref{eq:basic_model}. To see this, notice that the parameters of the non-structured prior in \eqref{eq:basic_model}, $\mu_a$ and $\Sigma_a$, are obtained by marginalizing out $\psi$ in \eqref{eq:contextual_gaussian_model}. In this case, $\mu_a = \W_a\mu$ and $\Sigma_a = \Sigma_a + \W_a\Sigma\W_a^\top$. The corresponding posterior covariance is $\hat{\Sigma}^{\textsc{ns}}_a = ((\Sigma_a + \W_a\Sigma\W_a^\top)^{-1} + G_{a})^{-1}$, and is generally larger than the covariance of \alg, $\hat{\Sigma}_a$ in \eqref{eq:action_posterior_gaussian}. This is more pronounced when the number of actions $K$ is large and when the latent parameters are more uncertain than the action parameters. Thus, the BSO bound of \alg is smaller due to the reduced posterior uncertainty it exhibits. Also, note that even when $\pi_*(X)$ is unobserved in the sample set $S$, \alg's posterior covariance $\hat{\Sigma}_{\pi_*(X)}$ can remain small since we use interactions with all actions to compute it. This contrasts with standard non-structured priors \eqref{eq:basic_model}, where observing $\pi_*(X)$ is necessary; without such observations, the posterior covariance $\hat{\Sigma}_{\pi_*(X)}$ would simply be the prior covariance $\Sigma_{\pi_*(X)}$.

Next, we provide a bound on the BSO that scales as \( \mathcal{O}(1/\sqrt{n}) \). To simplify the exposition, we roughly present its scaling with \( n \) in \cref{thm:main_thm} and defer the complete general statement to \cref{subsec:explicit_bound}.

\begin{theorem}[Scaling with $n$]\label{thm:main_thm}
For $n$ large enough and under some mild conditions stated in \cref{subsec:explicit_bound}, the BSO of \emph{\alg} under the structured prior \eqref{eq:contextual_gaussian_model} satisfies
\begin{align}
    \textsc{Bso}(\hat{\pi}_{\textsc{g}}) &= \mathcal{O}\Big(\sqrt{ d  \E{X\sim \nu}{e^{- \frac{n \pi_0^2(\pi_*(X)|X)}{2}}} + \frac{d}{n}} \nonumber\\ & \hspace{1cm}  + \sqrt{ \frac{d}{n} \E{X \sim \nu}{\frac{1}{ \pi_0(\pi_*(X)|X) + \frac{1}{n}}}}  \Big)\nonumber\,.
\end{align}
\end{theorem}

The above bound demonstrates \alg's scaling with \( n \) without relying on the standard \emph{well-explored dataset} assumption commonly used in the literature \citep{jin2021pessimism,hong2023multi}. For instance, \citet{hong2023multi} assumed that the dataset satisfies the condition \( G_a \succ \gamma n G \), where \( G = \mathbb{E}_{X \sim \nu}[\phi(X) \phi(X)^\top] \). Instead of such assumptions, we applied techniques from \citet{oliveira2016lower} and focused on high-probability events where these conditions naturally occur. This approach enables us to derive a scaling with \( n \) that directly links the constants to \( \pi_0 \), specifically the probability mass \( \pi_0 \) assigns to the optimal policy (\( \pi_0(\pi_*(X) \mid X) \)) averaged over contexts \( X \sim \nu \). As a result, the bound becomes smaller or larger depending on how well the logging policy \( \pi_0 \) covers the optimal actions for each context \( x \). For instance, one could replace the logging policy \( \pi_0 \) with a uniform distribution or the optimal policy, leading both to $\mathcal{O}(1/\sqrt{n})$ where the latter has a tighter bound independent of $K$.

\textbf{Technical Challenges.} Our proof is novel and presents several technical challenges for two main reasons. First, we bound a new metric, BSO, which prevents the use of standard frequentist techniques. Second, we avoid the common well-explored dataset assumption and instead provide a bound that directly depends on the logging policy \( \pi_0 \), making the dependence explicit rather than implicit in an assumption. Deriving this bound without such assumptions is challenging because we needed to control how often each action appears in the dataset. To address this, we modeled the number of appearances as a Binomial random variable with carefully chosen parameters and applied results from \citet{oliveira2016lower}, which provide high-probability bounds for our posterior covariance matrix. A sketch of the proof and an outline of these challenges are provided in \cref{subsec:explicit_bound}.

A similar result can be derive in OPE, where we bound the BMSE as follows.
\begin{theorem}[OPE Result]\label{thm:main_thm_2}
Let $x \in \cX$ and $a \in \cA$, the BMSE of \emph{\alg} under the structured prior \eqref{eq:contextual_gaussian_model} satisfies 
\begin{align*}
    {\textsc{Bmse}}(\hat{r}(x, a)) \leq \E{}{\|\phi(x)\|_{\hat{\Sigma}_{a}}^2}\,.
\end{align*}
\end{theorem}
For any context $x$ and action $a$, the BMSE of \alg is bounded by $ \mathbb{E}[\|\phi(x)\|_{\hat{\Sigma}_{a}}^2]$. This is expected, given that we assume access to well-specified prior and likelihood, eliminating bias and directly linking BMSE to the variance. Essentially, estimation accuracy increases as the posterior covariance of action $a$, $\hat{\Sigma}_a$, diminishes in the direction of $\phi(x)$. For the standard non-structured prior in \eqref{eq:basic_model}, this would only happen if the context $x$ and action $a$ appear frequently in the sample set $S$. However, \alg's use of structured priors ensures lower covariance in the direction of $\phi(x)$ even without observing context $x$ and action $a$ together. This is because \alg calculates posterior covariance for action $a$ using all observed contexts and actions, which reduces its variance. We will now present our main OPL result. 

\begin{figure*}
  \centering  \includegraphics[width=\linewidth]{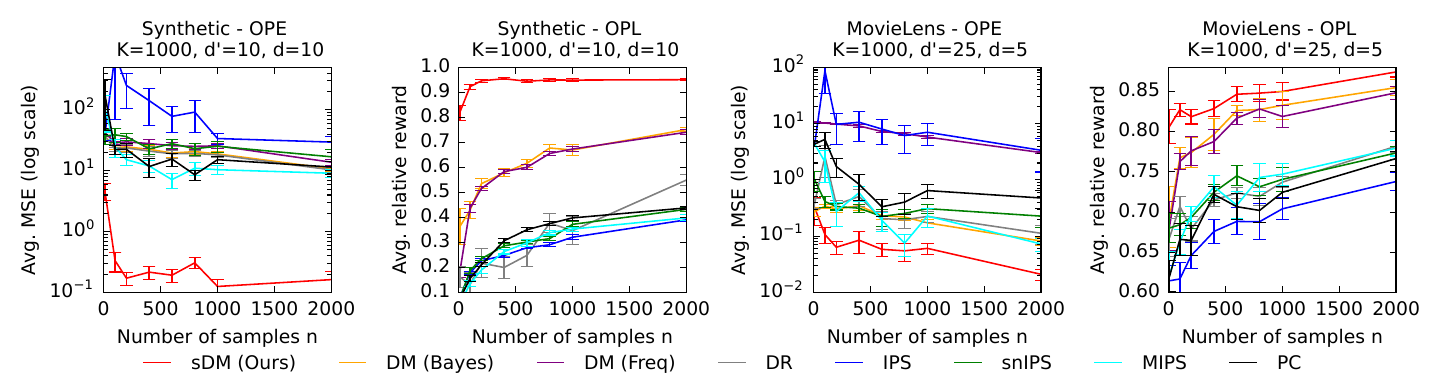}
  \caption{Performance of \alg and baselines on synthetic and MovieLens problems.} 
  \label{fig:synthetic_linear}
\end{figure*}

\section{EXPERIMENTS}\label{sec:experiments}
We evaluate \alg using both synthetic and real datasets. For OPL, we use the average reward relative to the optimal policy as the evaluation metric, and for OPE, we use the mean squared error (MSE). 

\subsection{Synthetic Problems}

\textbf{Setting.} We simulate synthetic data using the linear-Gaussian model in \eqref{eq:contextual_gaussian_model} with \( \sigma = 1 \). The contexts \( X \) are sampled uniformly from \( [-1, 1]^d \), with \( d = 10 \). The matrices \( \W_a \) are sampled uniformly from \( [-1,1]^{d \times d^\prime} \), where \( d^\prime = 10 \). We set \( \Sigma = 3I_{d^\prime} \) and \( \Sigma_a = I_d \), meaning the latent parameters are more uncertain than the action parameters. The latent mean \( \mu \) is randomly sampled from \( [-1, 1]^{d^\prime} \). The number of actions is \( K = 10^3 \), and we use a uniform logging policy to collect data. Additional experiments with different values of \( d' \), \( K \), \( d \), and logging policies are presented in \cref{app:experiments}.

\textbf{Baselines.} First, we use \alg under prior \eqref{eq:contextual_gaussian_model}. Second, we examine \texttt{DM (Bayes)}, which uses the standard non-structured prior \eqref{eq:basic_model}, where parameters $\mu_a$ and $\Sigma_a$ are obtained by marginalizing out the latent parameters $\psi$ in \eqref{eq:contextual_gaussian_model}. Thus \texttt{DM (Bayes)} is a standard Bayesian DM that does not capture arm reward correlations. We also include \texttt{DM (Freq)}, which estimates $\theta_{*, a}$ by the MLE. We include \texttt{\texttt{IPS}} \citep{horvitz1952generalization}, self-normalized \texttt{IPS} (\texttt{sn\texttt{IPS}}) \citep{swaminathan2015self}, and doubly robust (\texttt{DR}) \citep{dudik14doubly}, which we optimize to learn the optimal policy. M\texttt{IPS} \citep{saito2022off} and \texttt{PC} \citep{sachdeva2023off} are also included. Implementation details of baselines is provided in \cref{app:details_imp}.


\textbf{Results.} In \cref{fig:synthetic_linear} (first two columns), we plot the results for OPE and OPL. Overall, \alg consistently outperforms the baselines in both OPE and OPL. This performance gap becomes even more significant when sample size \(n\) is small. These results highlight \alg's enhanced efficiency in using available logged data, making it particularly beneficial in data-limited situations and scalable to large action spaces. \texttt{IPS} variants for large action spaces such as \texttt{MIPS} and \texttt{PC} outperform all other \texttt{IPS} variants and standard \texttt{DM} in OPE, but this increase in performance does not translate to OPL. 

\begin{figure}[H]
\begin{center}
\includegraphics[width=\linewidth]{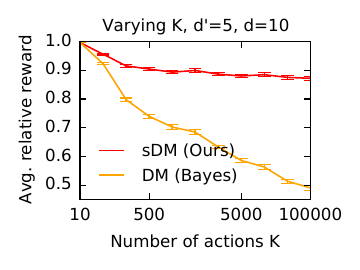}
\caption{\alg \emph{vs.} \texttt{DM (Bayes)} for varying $K$.}
\label{fig:las}
\end{center}
\end{figure}

\textbf{Scaling to large action spaces.} \texttt{sDM} achieves improved scalability compared to standard DM as leverages data more efficiently. While it still learns a $d$-dim. parameter for each action $a$, it does so by considering interactions with all actions in the sample set $S$, instead of only using interactions with the specific action $a$. This is crucial, especially given that many actions may not even be observed in $S$. To show \texttt{sDM}'s improved scalability, we compare it to the most competitive baseline in OPL simulations, \texttt{DM (Bayes)}, for varying $K \in [10,10^5]$ with $n=10^3$. The results in \cref{fig:las} reveal that the performance gap between \texttt{sDM} and \texttt{DM (Bayes)} becomes more significant when the number of actions $K$ increases. Hence, despite the necessity for \texttt{sDM} to learn distinct parameters for each action, accommodating practical scenarios like recommender systems where unique embeddings are learned for each product, it still enjoys good scalability.

\begin{figure}[H]
\begin{center}
\centerline{\includegraphics[width=\linewidth]{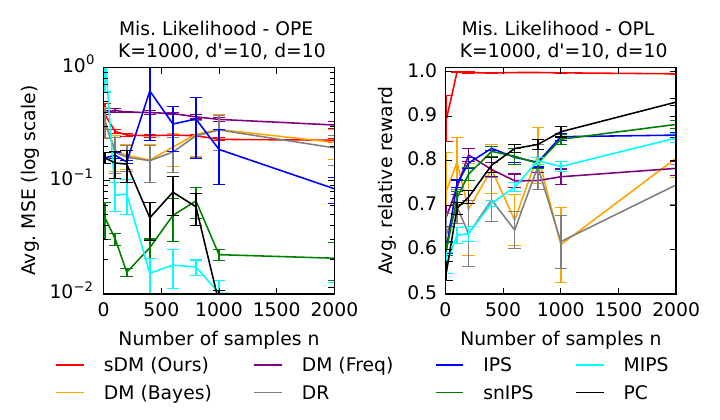}}
\caption{Effect of Misspecification.}
\label{fig:effect_likelihood_misspecfication}
\end{center}
\end{figure}

\textbf{Likelihood and Prior Misspecification.} We consider the case where \alg's assumed likelihood is misspecified. Specifically, we simulate binary rewards using a Bernoulli-logistic model, while \alg assumes a linear-Gaussian likelihood. Other DMs (\texttt{DM (Bayes)} and \texttt{DM (Freq)}) also use misspecified likelihoods. Results are shown in \cref{fig:effect_likelihood_misspecfication}. Despite the misspecification, \alg significantly outperforms all methods in OPL. However, \alg experiences a drop in OPE performance. This suggests that while the assumed Gaussian reward model does not predict the reward accurately, it still captures the ranking of actions. IPS variants like \texttt{MIPS}, \texttt{PC}, and \texttt{snIPS} perform well in OPE, but they fail to match (or even come close to) \alg's performance in OPL, even with misspecified likelihood. Additional experiments with misspecified priors, as presented in \cref{app:misspecification_experiments}, yield similar findings.

\subsection{MovieLens Problems}
\textbf{Setting.} We use MovieLens 1M \citep{movielens}, which contains 1 million ratings representing the interactions between 6,040 users and 3,952 movies. To create a semi-synthetic environment, we first apply a low-rank factorization to the rating matrix, producing 5-dim. representations: \( x_u \in \mathbb{R}^5 \) for user \( u \in [6040] \) and \( \theta_a \in \mathbb{R}^5 \) for movie \( a \in [3952] \). Movies are treated as actions, and contexts \( X \) are sampled randomly from the user vectors. The reward for movie \( a \) and user \( u \) is modeled as \( \mathcal{N}(x_u^\top \theta_a, 1) \), serving as proxy for ratings. A uniform logging policy is used to collect data.

\textbf{Baselines.} We consider the same baselines as in synthetic data. A prior is not needed for \texttt{DM (Freq)}, \texttt{\texttt{IPS}}, \texttt{sn\texttt{IPS}}, and \texttt{DR}. However, for \texttt{DM (Bayes)}, a standard prior \eqref{eq:basic_model} is inferred from data, where we set $\mu_a$ to be the mean of movie vectors across all dimensions, and $\Sigma_a = {\rm diag}(v)$, where $v$ represents the variance of movie vectors across all dimensions. Unlike the synthetic experiments, the latent structure assumed by \alg is not inherently present in MovieLens. But we learn it by training a Gaussian Mixture Model (GMM) to cluster movies into $J=5$ mixture components. This gives rise to the mixed-effect structure described in \cref{sec:application}, which represents a specific instance of \alg with $d^\prime = dJ = 25$. \texttt{MIPS} also has access to movie clusters, while we use the knn smoothing implementation of \texttt{PC} (see \citep[Section 3]{sachdeva2023off}). Note that \texttt{DM (Bayes)}, \alg, \texttt{MIPS} and \texttt{PC} use the same subset of data (of size $1000$) to learn their priors/assumed structure and thus we compare them fairly. We conduct experiments with $K = 10^3$ randomly selected movies. 

\textbf{Results.} Results are in \cref{fig:synthetic_linear} (last two columns). Even though the latent structure assumed by \alg is not inherently present in MovieLens, \alg still outperforms the baselines by learning it offline.

\subsection{KuaiRec Problems}

We also compare \alg to baselines using the KuaiRec dataset \citep{gao2022kuairec}, which is similar to MovieLens but with a key difference. Unlike MovieLens, which contains sparse user-movie interactions, KuaiRec offers a fully observed user-item interaction matrix. This allows us to evaluate algorithms without relying on semi-synthetic simulations based on low-rank factorization models to complete the matrix. This setup enables a comparison of \alg with baselines in a setting where neither the prior nor the likelihood is well-specified for \alg (and other DMs). The true reward model is unknown and not necessarily Gaussian, and no simulated environment is sampled from a prior. Due to space constraints, details on data preprocessing and filtering of KuaiRec, along with additional OPE results, are provided in \cref{app:kuairec}.

\begin{figure}
\begin{center}
\includegraphics[width=\linewidth]{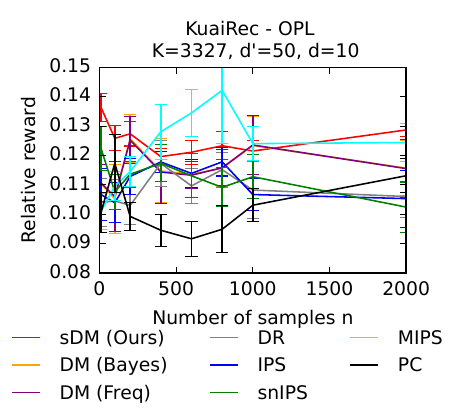}
\caption{OPL Results on KuaiRec.}
\label{fig:kuairec}
\end{center}
\end{figure}

The results are shown in \cref{fig:kuairec}. First, \alg outperforms all DM variants, including \texttt{DM (Bayes)} and \texttt{DM (Freq)}, making it the most competitive DM baseline in these experiments. When compared to IPS variants, \alg outperforms all of them except for \texttt{MIPS}, where the performance comparison depends on the sample size \( n \). In low-data regimes (small $n$), \alg outperforms \texttt{MIPS}, whereas \texttt{MIPS} outperforms \alg when the number of samples is sufficiently large. \alg slightly outperforms \texttt{MIPS} when $n=2000$.

\section{CONCLUSION}\label{sec:conclusion}
We introduced \alg, a novel approach for OPE and OPL that leverages latent structures among actions to enhance statistical efficiency while maintaining computational feasibility, particularly in scenarios with large action spaces and limited data. We developed a Bayesian framework to assess performance, showing the superiority of greedy policies over pessimistic ones and providing theoretical justifications for \alg's advantages. However, a key limitation of our theory is the assumption of a well-specified prior. While we have empirically studied the effects of misspecification, a theoretical investigation remains for future work. Additionally, extending \alg to handle non-linear hierarchies is a promising direction for further exploration.

\subsubsection*{Acknowledgments}
We sincerely thank the anonymous reviewers for their valuable feedback and constructive suggestions, which have helped improve the clarity and quality of this work. We are also grateful to Hugo Chardon for pointing us to key references that enabled us to bound the covariance matrix under less restrictive conditions.

\bibliography{biblio}
\bibliographystyle{plainnat}

\section*{Checklist}

 \begin{enumerate}

 \item For all models and algorithms presented, check if you include:
 \begin{enumerate}
   \item A clear description of the mathematical setting, assumptions, algorithm, and/or model. [Yes]
   \item An analysis of the properties and complexity (time, space, sample size) of any algorithm. [Yes]
   \item (Optional) Anonymized source code, with specification of all dependencies, including external libraries. [Yes]
 \end{enumerate}

 \item For any theoretical claim, check if you include:
 \begin{enumerate}
   \item Statements of the full set of assumptions of all theoretical results. [Yes]
   \item Complete proofs of all theoretical results. [Yes]
   \item Clear explanations of any assumptions. [Yes]     
 \end{enumerate}

 \item For all figures and tables that present empirical results, check if you include:
 \begin{enumerate}
   \item The code, data, and instructions needed to reproduce the main experimental results (either in the supplemental material or as a URL). [Yes]
   \item All the training details (e.g., data splits, hyperparameters, how they were chosen). [Yes]
         \item A clear definition of the specific measure or statistics and error bars (e.g., with respect to the random seed after running experiments multiple times). [Yes]
         \item A description of the computing infrastructure used. (e.g., type of GPUs, internal cluster, or cloud provider). [Yes]
 \end{enumerate}

 \item If you are using existing assets (e.g., code, data, models) or curating/releasing new assets, check if you include:
 \begin{enumerate}
   \item Citations of the creator If your work uses existing assets. [Yes]
   \item The license information of the assets, if applicable. [Yes]
   \item New assets either in the supplemental material or as a URL, if applicable. [Yes]
   \item Information about consent from data providers/curators. [Yes]
   \item Discussion of sensible content if applicable, e.g., personally identifiable information or offensive content. [Not Applicable]
 \end{enumerate}

 \item If you used crowdsourcing or conducted research with human subjects, check if you include:
 \begin{enumerate}
   \item The full text of instructions given to participants and screenshots. [Not Applicable]
   \item Descriptions of potential participant risks, with links to Institutional Review Board (IRB) approvals if applicable. [Not Applicable]
   \item The estimated hourly wage paid to participants and the total amount spent on participant compensation. [Not Applicable]
 \end{enumerate}

 \end{enumerate}

\newpage
\appendix
\onecolumn

\section*{ORGANIZATION OF THE SUPPLEMENTARY MATERIAL}
The supplementary material is organized as follows. 
\begin{itemize}
    \item In \textbf{\cref{app:preliminaries}}, we provide a detailed notation.
    \item In \textbf{\cref{sec:extended_related_work}}, we provide an extended related work discussion.
    \item In \textbf{\cref{app:posterior_derivations_standard}}, we outline the posterior derivations for the standard prior in \eqref{eq:basic_model}.  
    \item In \textbf{\cref{app:posterior_derivations}}, we outline the posterior derivations under the structured prior in \eqref{eq:contextual_gaussian_model}.
    \item In \textbf{\cref{sec:ope_proofs}}, we prove the claims made in \cref{sec:analysis}.
    \item In \textbf{\cref{app:experiments}}, we present supplementary experiments.
    \item In \textbf{\cref{app:impact,app:compute}}, we discuss the broader impact of this work and the moderate amount of computing needed.

\end{itemize}

\section{DETAILED NOTATION}\label{app:preliminaries}

For any positive integer $n$, we define $[n]$ as the set ${1,2,\ldots,n}$. We represent the identity matrix of dimension $d \times d$ as $I_d$. In our notation, unless explicitly stated otherwise, the $i$-th coordinate of a vector $v$ is denoted as $v_i$. However, if the vector is already indexed, such as $x_i$, we use the notation $x_{i, j}$ to represent the $j$-th entry of the vector $x_i$. When dealing with matrices, we refer to the $(i, j)$-th entry of a matrix $\Alpha$ as $\Alpha_{i, j}$. Also, $\lambda_1(\Alpha)$ and $\lambda_d(\Alpha)$ refer to the maximum and minimum eigenvalues of matrix $\Alpha$, respectively. Moreover, for any positive-definite matrix $A$ and vector $x$, we let $\| x \|_A = \sqrt{x^\top A x}$.

Now, consider a collection of $n$ vectors, denoted as $a_1 \in \real^d, a_2 \in \real^d, \ldots, a_n \in \real^d$. Then we use $a = (a_i)_{i \in [n]} \in \mathbb{R}^{nd}$ to represent a $nd$-dimensional vector formed by concatenating the vectors $a_1, a_2, \ldots, a_n$. The operator $\operatorname{Vec}(\cdot)$ is used to vectorize a matrix or a set of vectors. Let $\Alpha_1, \Alpha_2, \ldots, \Alpha_n$ be a collection of $n$ matrices, each of dimension $d \times d$. The notation $\mathrm{diag}((\Alpha_i)_{i \in [n]}) \in \real^{nd \times nd}$ represents a block diagonal matrix where $\Alpha_1, \Alpha_2, \ldots, \Alpha_n$ are the main-diagonal blocks. Similarly, $(\Alpha_i)_{i \in [n]} \in \real^{nd \times d}$ denotes the $nd \times d$ matrix formed by concatenating matrices $\Alpha_1, \Alpha_2, \ldots, \Alpha_n$. To provide a more visual representation, consider a collection of vectors $a_1 \in \real^d, a_2 \in \real^d, \ldots, a_n \in \real^d$ and matrices $\Alpha_1 \in \real^{d \times d}, \Alpha_2 \in \real^{d \times d}, \ldots, \Alpha_n \in \real^{d \times d}$. We can represent them as follows
\begin{align*}
    &(a_i)_{i \in [n]} = {\begin{pmatrix} a_1 \\ a_2 \\ \vdots \\ a_n \end{pmatrix}} \in \real^{nd}\,, \quad
    \mathrm{diag}((\Alpha_i)_{i \in [n]}) = {\begin{pmatrix}\Alpha _{1}&0&\cdots &0\\0&\Alpha _{2}&\cdots &0\\\vdots &\vdots &\ddots &\vdots \\0&0&\cdots &\Alpha _{n}\end{pmatrix}} \in \real^{nd \times nd}\,, \\ 
    &(\Alpha_i)_{i \in [n]} = {\begin{pmatrix}  \Alpha_1 \\ \Alpha_2 \\ \vdots \\ \Alpha_n \end{pmatrix}} \in \real^{nd \times d}\,.
\end{align*}

\section{EXTENDED RELATED WORK}\label{sec:extended_related_work}
Online learning under uncertainty is often modeled within the framework of contextual bandits \citep{lattimore19bandit,li10contextual,chu11contextual}. Naturally, learning within this framework follows an online paradigm. However, practical applications often entail a large action space and a significant emphasis on short-term gains. This presents a challenge where the agent needs to exhibit a \textit{risk-averse} behavior, contradicting the foundational principle of online algorithms designed to explore actions for long-term benefit \citep{auer02finitetime,thompson33likelihood,russo18tutorial,agrawal13thompson,abbasi-yadkori11improved}. While several practical online algorithms have emerged to efficiently explore the action space in contextual bandits \citep{zong2016cascading,zhu2022contextual,aoualimixed,aouali2023linear}, a notable gap exists in the quest for an offline procedure capable of optimizing decision-making using historical data. Fortunately, we are often equipped with a sample set collected from past interactions. Leveraging this data, agents can enhance their policies offline \citep{swaminathan2015batch,london2019bayesian,sakhi2022pac}, consequently improving the overall system performance. This study is primarily concerned with this offline or \textit{off-policy} formulation of contextual bandits \citep{dudik2011doubly,dudik2012sample,dudik14doubly,wang2017optimal,farajtabar2018more}. There are two main tasks in off-policy contextual bandits, first, off-policy evaluation (OPE) \citep{dudik2011doubly} which involves estimating policy performance using historical data, simulating as if these evaluations were performed while the policy is interacting with the environment in real-time. Subsequently, the derived estimator is refined to closely approximate the optimal policy, a process known as off-policy learning (OPL) \citep{swaminathan2015batch}. Next, we review both OPE and OPL.

\subsection{Off-Policy Evaluation}
\label{sec:related_work_OPE}
Recent years have witnessed a surge in interest in OPE, with several key works contributing to this field \citep{dudik2011doubly,dudik2012sample,dudik14doubly,wang2017optimal,farajtabar2018more,su2019cab,su2020doubly,metelli2021subgaussian,kuzborskij2021confident,saito2022off,sakhi2020blob,jeunen2021pessimistic}. The literature on OPE can be broadly categorized into three main approaches. The first approach referred to as the direct method (DM) \citep{jeunen2021pessimistic,hong2023multi}, involves learning a model that approximates the expected reward. This model is then used to estimate the performance of evaluated policies. The second approach, inverse propensity scoring (IPS) \citep{horvitz1952generalization,dudik2012sample}, aims to estimate the reward of evaluated policies by correcting for the preference bias of the logging policy in the sample set. IPS is unbiased when there's an assumption that the evaluation policy is absolutely continuous with respect to the logging policy. However, it can exhibit high variance and substantial bias when this assumption is violated \citep{sachdeva2020off}. Various techniques have been introduced to address the variance issue, such as clipping importance weights \citep{ionides2008truncated,swaminathan2015batch}, smoothing them \citep{aouali2023exponential}, self-normalization of weights \citep{swaminathan2015self}, etc. \citep{gilotte2018offline}. The third approach, known as doubly robust (DR) \citep{robins1995semiparametric,bang2005doubly,dudik2011doubly,dudik14doubly,farajtabar2018more}, combines elements of DM and IPS, helping to reduce variance. Assessing the accuracy of an OPE estimator, denoted as $\hat{R}_{n}(\pi)$, is typically done using the mean squared error (MSE). It may be relevant to note that \citet{metelli2021subgaussian} argued that high-probability concentration rates should be preferred over the MSE to evaluate OPE estimators as they provide non-asymptotic guarantees. In this work, we presented a direct method for OPE, for which we could derive both an MSE and high-probability concentration bound under our assumptions. However, we focused more on OPL.

\subsection{Off-Policy Learning}
\label{sec:opl_learning_principles}
\textbf{OPL under IPS.} Prior research on OPL predominantly revolved around developing learning principles influenced by generalization bounds for IPS. \citet{swaminathan2015batch} proposed to penalize the IPS estimator with a variance term leading to a learning principle that promotes policies that achieve high estimated reward and exhibit minimal empirical variance. In a recent development, London et al. \citep{london2019bayesian} made a connection between PAC-Bayes theory and OPL that inspired several works in the same vein \citep{sakhi2022pac,aouali2023exponential}. \citet{london2019bayesian} proposed a learning principle that favors policies with high estimated reward while keeping the parameter close to that of the logging policy in terms of $L_2$ distance. In particular, this learning principle is scalable. Then \citet{sakhi2022pac,aouali2023exponential} took a different direction of deriving tractable generalization bounds and optimizing them directly as they are. Finally, \citet{wang2023oracle} recently introduced an efficient learning principle with guarantees for a specific choice of their hyperparameter. All of these learning principles are related to the concept of pessimism which we discuss next.

\textbf{OPL under DM.} The majority of OPL approaches in contextual bandits rely on the principle of pessimism \citep{jeunen2021pessimistic,jin2021pessimism,hong2023multi}. In essence, these methods construct confidence intervals for reward estimates  $\hat{r}(x, a),$ that hold simultaneously for all $(x, a) \in \cX \times \cA$ and satisfy the following condition
\begin{align*}
    |r(x, a; \theta) - \hat{r}(x, a)| & \leq u(x, a)\,, & \forall (x, a) \in \cX \times \cA
\end{align*}
Subsequently, the policy learned with this pessimistic approach is defined as
\begin{align}
     \hat{\pi}_{\textsc{p}}(a \mid x) = \mathds{1}{\{a = \argmax_{b \in \cA} \hat{r}(x, b) - u(x, b)\}}\,.
\end{align}
Note that when the function $u(x, a)$ is independent of both $x$ and $a$, the pessimistic and greedy policies become equivalent. However, in this work, we highlight the critical role of the selected metric and the underlying assumptions about $\hat{r}$ in choosing between pessimistic and greedy policies. In particular, we introduce the concept of Bayesian suboptimality (BSO), as defined in  \cref{subsec:bayes_suboptimality}. BSO evaluates the average performance of algorithms across multiple problem instances, and in this context, the Greedy policy emerges as a more suitable choice for minimizing the BSO. A detailed comparison of pessimistic and greedy policies is presented in \cref{subsec:greedy_pessimism}.

In addition to not using pessimism, our work also considers structured contextual bandit problems, upon which we build informative priors. Notably, while structured problems have been previously investigated in the context of multi-task learning by \citet{lazaric2010bayesian,hong2023multi}, these works primarily focused on cases where there are multiple contextual bandit instances that bear similarity to one another. Interaction with one bandit instance contributes to the agent's understanding of other instances. In contrast, our work tackles a single contextual bandit instance with a large action space and considers an inherent structure among actions within this bandit instance. Furthermore, the structure we address in the single-instance problem is different and more general compared to the ones explored in the aforementioned multi-task learning works.

\section{POSTERIOR DERIVATIONS UNDER STANDARD PRIORS}\label{app:posterior_derivations_standard}
Here we derive the posterior under the standard prior in \eqref{eq:basic_model}. These are standard derivations and we present them here for the sake of completeness. But first, we state the following standard assumption that allows posterior derivations. 
\begin{assumption}[Independence]\label{ass:independence}
 $(X, A)$ is independent of $\theta$, and the $\theta_a$, for $a \in \cA$ are independent. 
\end{assumption}

\begin{proof}[Derivation of 
 $\, p(\theta_a \mid S)$ for the standard prior in \eqref{eq:basic_model}] We start by recalling the standard prior in \eqref{eq:basic_model}
\begin{align}\label{eq:app_basic_model}
  \theta_a &\sim \cN(\mu_{a}, \Sigma_{a})\,, &  \forall a \in \cA\,, \\
    R \mid \theta, X, A &\sim \cN(\phi(X)^\top \theta_A , \sigma^2)\,,& \nonumber
\end{align}
where $\cN(\mu_{a}, \Sigma_{a})$ is the prior on the action parameter $\theta_a$. Let $\theta = (\theta_a)_{a \in \cA} \in \real^{dK}$, $\Sigma_{\cA} = {\rm diag}(\Sigma_a)_{a \in \cA} \in \real^{dK \times dK}$ and $\mu_{\cA} = (\mu_a)_{a\in \cA} \in \real^{dK}$. Also, let $u_a \in \set{0, 1}^K$ be the binary vector representing the action $a$. That is, $u_{a, a} = 1$ and $u_{a, a^\prime} = 0$ for all $a^\prime \neq a$. Then we can rewrite the model in \eqref{eq:app_basic_model} as 
\begin{align}\label{eq:app_joint_basic_model}
  \theta &\sim \cN(\mu_{\cA}, \Sigma_{\cA})\,, \\
    R \mid \theta, X, A &\sim \cN((u_A \otimes \phi(X))^\top \theta , \sigma^2)\,. \nonumber
\end{align}
Then the \emph{joint} action posterior $p(\theta \mid S)$ decomposes as
\begin{align*}
    p(&\theta \mid S) = p(\theta \mid (X_i, A_i, R_i)_{i \in [n]}) \stackrel{(i)}{\propto} p((R_i)_{i \in [n]} \mid \theta, (X_i, A_i)_{i \in [n]}) p(\theta \mid  (X_i, A_i)_{i \in [n]}) \,,  \\
    &\stackrel{(ii)}{=} p((R_i)_{i \in [n]} \mid \theta, (X_i, A_i)_{i \in [n]}) p(\theta) \stackrel{(iii)}{=} \prod_{i \in [n]}p(R_i \mid \theta, X_i, A_i) p(\theta) \,,  \\
    & \stackrel{(iv)}{=}  \prod_{i \in [n]} \cN(R_i; (u_{A_i} \otimes  \phi(X_i))^\top \theta, \sigma^2) \cN(\theta; \mu_{\cA} , \Sigma_{\cA})\,,\\
    & = \exp\left[- \frac{1}{2}
  \left(v \sum_{i =1}^n (R_i^2 - 2 R_i (u_{A_i} \otimes \phi(X_i))^\top \theta + ((u_{A_i} \otimes \phi(X_i))^\top \theta)^2) + \theta^\top \Lambda_{\cA} \theta - 2 \theta^\top \Lambda_{\cA} \mu_{\cA} + \mu_{\cA}^\top \Lambda_{\cA} \mu_{\cA}\right)\right] \,,\\
  &\propto \exp\left[- \frac{1}{2}
  \left( \theta^\top(v\sum_{i =1}^n (u_{A_i} \otimes \phi(X_i)) (u_{A_i} \otimes \phi(X_i))^\top + \Lambda_{\cA}) \theta - 2 \theta^\top \left(v \sum_{i =1}^n (u_{A_i} \otimes \phi(X_i))^\top R_i + \Lambda_{\cA} \mu_{\cA} \right)\right)\right]\,,\\
    &= \exp\left[- \frac{1}{2}
  \left( \theta^\top(v\sum_{i =1}^n (u_{A_i}u_{A_i}^\top \otimes \phi(X_i) \phi(X_i)^\top)+ \Lambda_{\cA}) \theta - 2 \theta^\top \left(v \sum_{i =1}^n (u_{A_i} \otimes \phi(X_i))^\top R_i + \Lambda_{\cA} \mu_{\cA} \right)\right)\right]\,,\\
  &\stackrel{(v)}{\propto} \cN\left(\theta; \hat{\mu}_{\cA}, \left(\hat{\Lambda}_{\cA}\right)^{-1}\right)\,.
\end{align*}
In $(i)$, we apply Bayes rule. In $(ii)$, we use that $\theta$ is independent of $(X, A)$, and $(iii)$ follows from the assumption that $R_i \mid \theta, X_i, A_i$ are i.i.d. Finally, in $(iv)$, we replace the distribution by their Gaussian form, and in $(v)$, we set $\hat{\Lambda}_{\cA} = v\sum_{i =1}^n (u_{A_i}u_{A_i}^\top \otimes \phi(X_i) \phi(X_i)^\top) + \Lambda_{\cA}\,,$ and $  \hat{\mu}_a = \hat{\Lambda}_{\cA}^{-1} \big( v \sum_{i =1}^n (u_{A_i} \otimes \phi(X_i))^\top R_i + \Lambda_{\cA} \mu_{\cA}\Big)$. Now notice that $\hat{\Lambda}_{\cA} = {\rm diag}(\Sigma_a^{-1} + G_a)_{a \in \cA}$. Thus, $p(\theta \mid S) = \cN(\theta, \hat{\mu}_{\cA}, \Lambda_{\cA}^{-1})$ where $\hat{\mu}_{\cA} = (\hat{\mu}_{a})_{a \in \cA}$ and $ \Lambda_{\cA} = {\rm diag}(\hat{\Lambda}_{a})_{a \in \cA}\,,$ with
\begin{align*}
    &\hat{\Lambda}_{a} = \Sigma_a^{-1} + G_a\,, &\hat{\Lambda}_{a}\hat{\mu}_{a} &=  \Sigma_a^{-1} \mu_a + B_a \,.
\end{align*}
Since the covariance matrix of $p(\theta \mid S)$ is diagonal by block, we know that the marginals $\theta_a \mid S$ also have a Gaussian density  $p(\theta_a \mid S) = \cN(\theta_a; \hat{\mu}_{a}, \hat{\Sigma}_{a})$ where $\hat{\Sigma}_{a} = \hat{\Lambda}_{a}^{-1}$.

\end{proof}

\section{POSTERIOR DERIVATIONS UNDER STRUCTURED PRIORS}\label{app:posterior_derivations}
Here we derive the posteriors under the structured prior in \eqref{eq:contextual_gaussian_model}. Precisely, we derive the latent posterior density of $\psi \mid S$, the conditional posterior density of $\theta \mid S, \psi$. Then, we derive the marginal posterior $\theta \mid S$. Posterior derivations rely on the following assumption. 

\begin{assumption}[Structured Independence]\label{ass:indep_assumptions} \textit{(i)} $(X, A)$ is independent of $\psi$ and given $\psi$, $(X, A)$ is independent of $\theta$. \textit{(ii)} Given $\psi$, the $\theta_a,$ for all $a \in \cA$ are independent.  \end{assumption}

\subsection{Latent Posterior}\label{app:latent posterior-derivation}

\begin{proof}[Derivation of 
 $\, p(\psi \mid S)$]
First, recall that our model in \eqref{eq:contextual_gaussian_model} reads
\begin{align}
    \psi & \sim \cN(\mu, \Sigma)\,, \nonumber\\ 
    \theta_a \mid  \psi& \sim \cN\Big( \W_a \psi , \,  \Sigma_{a}\Big)\,, & \forall a \in \cA\,,\nonumber\\ 
    R  \mid \psi, \theta, X, A & \sim \cN(\phi(X)^\top \theta_A ,  \sigma^2)\,.
\end{align}
Then we first rewrite it as
\begin{align}\label{eq:model_rewritten_joint}
    \psi & \sim \cN(\mu, \Sigma)\,, \nonumber \\ 
    \theta \mid  \psi& \sim \cN\Big( \W_{\cA} \psi , \,  \Sigma_{\cA}\Big)\,,\nonumber\\ 
   R  \mid \psi, \theta, X, A  & \sim \cN((u_A \otimes \phi(X))^\top \theta ,  \sigma^2)\,.
\end{align}
Then the latent posterior is
\begin{align*}
   p(\psi \mid (X_i, A_i, R_i)_{i \in [n]})  & \propto p((R_i)_{i \in [n]} \mid \psi, (X_i, A_i)_{i \in [n]})p(\psi \mid (X_i, A_i)_{i \in [n]}) \,,\\
    &\stackrel{(i)}{=} p((R_i)_{i \in [n]} \mid \psi, (X_i, A_i)_{i \in [n]}) q(\psi)\,,\\
    &= \int_{\theta} p((R_i)_{i \in [n]}, \theta \mid \psi, (X_i, A_i)_{i \in [n]}) \dif \theta \, q(\psi)\,,\\
    &= \int_{\theta} p((R_i)_{i \in [n]} \mid \psi, \theta, (X_i, A_i)_{i \in [n]}) p(\theta \mid \psi, (X_i, A_i)_{i \in [n]}) \dif \theta \, q(\psi)\,,\\
    &\stackrel{(ii)}{=} \int_{\theta} p((R_i)_{i \in [n]} \mid \psi, \theta, (X_i, A_i)_{i \in [n]}) p(\theta \mid \psi) \dif \theta \, q(\psi)\,,\\
\end{align*}
In $(i)$, we use that $(X, A)$ is independent of $\psi$, which follows from \cref{ass:indep_assumptions}. Similarly, in $(ii)$, we use that $\theta$ is conditionally independent of $(X, A)$ given $\psi$. Now we know that given $\theta$, $R_i \mid X_i, A_i$ are i.i.d. and hence $p((R_i)_{i \in [n]} \mid \psi, \theta, (X_i, A_i)_{i \in [n]}) = \prod_{a \in \cA} \LL_a(\theta_a)$. Moreover, $\theta_a$ for $a \in \cA$ are conditionally independent given $\psi$. Thus $p(\theta  \mid \psi) = \prod_{a \in \cA} p_a\left(\theta_a; f_a(\psi) \right)$, where we also used that $\theta_a \mid \psi \sim  p_a\left(\cdot; f_a(\psi) \right)$. This leads to 
\begin{align*}
     p(\psi \mid (X_i, A_i, R_i)_{i \in [n]})  & \propto \int_{\theta}  \prod_{a \in \cA} \LL_a(\theta_a) p_a\left(\theta_a ; f_a(\psi) \right) \dif \theta \, q(\psi)\,,\\
    &\stackrel{(i)}{=} \prod_{a \in \cA} \int_{\theta_a} \LL_a(\theta_a) \cN \left(\theta_a ; \W_a \psi, \Sigma_a\right) \dif \theta_a \cN(\psi;\mu, \Sigma)\,, \nonumber\\
   & \stackrel{(ii)}{=} \prod_{a \in \cA} \int_{\theta_a}  \Big(\prod_{i \in I_a} \cN(R_i; \phi(X_i)^\top \theta_a, \sigma^2)\Big) \cN \left(\theta_a ; \W_a \psi, \Sigma_a\right) \dif \theta_a \cN(\psi;\mu, \Sigma)\,.
\end{align*}
In $(i)$, we notice that $\theta=(\theta_a)_{a \in \cA}$ and apply Fubini's Theorem. In $(ii)$, we let $I_a = |S_a| = \sum_{i \in [n]}\mathds{1}{\{A_i=a\}}$ as the number of times action $a$ appears in the sample set $S$.   Now let $h_a(\psi) = \int_{\theta_a} \left(\prod_{i \in I_a} \cN(R_i; \phi(X_i)^\top \theta_a, \sigma^2)\right) \cN \left(\theta_a; \W_a \psi, \Sigma_a\right) \dif \theta$. Then we have that 
\begin{align}\label{eq:helper}
    p(\psi \mid S)  & \propto \prod_{a \in \cA} h_a(\psi) \cN(\psi;\mu, \Sigma)\,.
\end{align}
We start by computing $h_a$. To reduce clutter, let $v = \sigma^{-2}$ and $\Lambda_a = \Sigma_a^{-1}$. Then we compute $h_a$ as 
\begin{align*}
& h_a(\psi)  =  \int_{\theta_a} \left(\prod_{i \in I_a} \cN(R_i; \phi(X_i)^\top \theta_a, \sigma^2)\right)
  \cN(\theta_a; \W_a \psi, \Sigma_a) \dif\theta_a\,, \\
  & \propto \int_{\theta_a} \exp\left[
  - \frac{1}{2} v \sum_{i \in I_a} (R_i - \phi(X_i)^\top \theta_a)^2 -
  \frac{1}{2} (\theta_a- \W_a \psi)^\top \Lambda_a (\theta_a- \W_a \psi)\right] \dif \theta_a\,,\\
  & = \int_{\theta_a} \exp\Big[- \frac{1}{2}
  \Big(v \sum_{i \in I_a} (R_i^2 - 2 R_i \theta_a^\top \phi(X_i) + (\theta_a^\top \phi(X_i))^2) +
  \theta_a^\top \Lambda_a \theta_a- 2 \theta_a^\top \Lambda_a \W_a \psi \\& \hspace{12cm} + (\W_a \psi)^\top \Lambda_a (\W_a \psi) \Big)\Big] \dif \theta_a\,,\\
  & \propto \int_{\theta_a} \exp\Big[- \frac{1}{2}
  \Big(\theta_a^\top \left( v \sum_{i \in I_a} \phi(X_i) \phi(X_i)^\top + \Lambda_a \right) \theta_a- 2 \theta_a^\top \left(v \sum_{i \in I_a} R_i \phi(X_i) + \Lambda_a \W_a \psi\right) \\ & \hspace{12cm} + (\W_a \psi)^\top \Lambda_a (\W_a \psi)\Big) \Big] \dif \theta_a\,.
 \end{align*}
Now recall that $G_a = v \sum_{i \in I_a} \phi(X_i) \phi(X_i)^\top$ and $B_a = v \sum_{i \in I_a} R_i \phi(X_i)$ and let $V_a = \left(G_a + \Lambda_a\right)^{-1}$, $U_a = V_a^{-1}$, and $\beta_a = V_a (B_a  + \Lambda_a \W_a \psi)$. Then have that $U_a V_a = V_a U_a = I_{d}\,,$ and thus
 \begin{align*}
  f(\psi) & \propto \int_{\theta_a} \exp\left[- \frac{1}{2}
  \left(\theta_a^\top U_a \theta_a- 2 \theta_a^\top U_a V_a\left(B_a + \Lambda_a \W_a \psi\right) + (\W_a \psi)^\top \Lambda_a (\W_a \psi)\right) \right] \dif \theta_a\,,\\
  & = \int_{\theta_a} \exp\left[- \frac{1}{2}
  \left(\theta_a^\top U_a \theta_a- 2 \theta_a^\top U_a \beta_a +  (\W_a \psi)^\top \Lambda_a (\W_a \psi)\right) \right] \dif \theta_a\,,\\
  & = \int_{\theta_a} \exp\left[- \frac{1}{2}
  \left( (\theta_a- \beta_a)^\top U_a (\theta_a- \beta_a) - \beta_a^\top U_a \beta_a + (\W_a \psi)^\top \Lambda_a (\W_a \psi)\right) \right] \dif \theta_a\,, \\
  & \propto  \exp\left[- \frac{1}{2}
  \left( - \beta_a^\top U_a \beta_a + (\W_a \psi)^\top \Lambda_a (\W_a \psi)\right) \right]\,,\\
  & =  \exp\left[- \frac{1}{2}
  \left( -  \left(B_a + \Lambda_a \W_a \psi\right)^\top  V_a \left(B_a + \Lambda_a \W_a \psi\right) + (\W_a \psi)^\top \Lambda_a (\W_a \psi)\right) \right] \,,\\
  & \propto  \exp\left[- \frac{1}{2}
  \left( \psi^{\top} \W_a^{\top} \left( \Lambda_a - \Lambda_a V_a \Lambda_a \right) \W_a \psi - 2 \psi^\top \left(\W_a^\top \Lambda_a V_a B_a\right) \right) \right] \,,\\
  & \propto \cN(\psi; \bar{\mu}_{a}, \bar{\Sigma}_{a})\,,
\end{align*}
where 
\begin{align}\label{eq:latent_posterior_formulas}
     \bar{\Sigma}_{a}^{-1} & = \W_a^{\top} \left( \Lambda_a - \Lambda_a V_a \Lambda_a \right)\W_a = \W_a^{\top} \left( \Sigma_a^{-1} - \Sigma_a^{-1} (G_a + \Sigma_a^{-1})^{-1} \Sigma_a^{-1} \right)\W_a\,, \nonumber \\
    \bar{\Sigma}_{a}^{-1} \bar{\mu}_{a} & =  \W_a^\top \Lambda_a V_a B_a =   \W_a^\top \Sigma_a^{-1} (G_a + \Sigma_a^{-1})^{-1}  B_a\,.
\end{align} 
However, we know from \eqref{eq:helper} that $p(\psi \mid S)  \propto \prod_{a \in \cA} h_a(\psi) \cN(\psi;\mu, \Sigma)$. But $h_a(\psi)$ is proportional to $\cN(\psi; \bar{\mu}_{a}, \bar{\Sigma}_{a})$ for any $a$. Thus $ p(\psi \mid S)$ can be seen as the product of $K+1$ multivariate Gaussian distributions $\cN(\mu, \Sigma)$ and $\cN(\bar{\mu}_{a}, \bar{\Sigma}_{a})$ for $a \in \cA$ (the terms $h_a$). Thus, $ p(\psi \mid S)$ is also a multivariate Gaussian distribution $\cN(\bar{\mu}, \bar{\Sigma}^{-1})$, with
\begin{align}\label{eq:without_woodbury}
    \bar{\Sigma}^{-1} &=  \Sigma^{-1} + \sum_{a \in \cA}\bar{\Sigma}_{a}^{-1} = \Sigma^{-1} + \sum_{a \in \cA} \W_a^{\top} \left( \Sigma_a^{-1} - \Sigma_a^{-1} (G_a + \Sigma_a^{-1})^{-1} \Sigma_a^{-1} \right)\W_a\,, \\
   \bar{\Sigma}^{-1}\bar{\mu} &=  \Sigma^{-1} \mu +\sum_{a \in \cA}\bar{\Sigma}_{a}^{-1}\bar{\mu}_{a} = \Sigma^{-1} \mu +\sum_{a \in \cA}   \W_a^\top \Sigma_a^{-1} (G_a + \Sigma_a^{-1})^{-1}  B_a\,.
\end{align}
\end{proof}

\subsection{Conditional Posterior}\label{app:conditional-posterior-derivation}

\begin{proof}[Derivation of 
 $\, p(\theta_a \mid \psi, S)$]
Let $v
  = \sigma^{-2}\,, \quad
  \Lambda_a
  = \Sigma_a^{-1}\,.$ We consider the model rewritten in \eqref{eq:model_rewritten_joint},  then the \emph{joint} action posterior $p(\theta \mid S)$ decomposes as
\begin{align*}
   & p(\theta \mid \psi,  S) = p(\theta \mid \psi, (X_i, A_i, R_i)_{i \in [n]}) \stackrel{(i)}{\propto} p((R_i)_{i \in [n]} \mid \theta,  \psi, (X_i, A_i)_{i \in [n]}) p(\theta \mid \psi,  (X_i, A_i)_{i \in [n]}) \,,  \\
    &\stackrel{(ii)}{=} p((R_i)_{i \in [n]} \mid \theta, (X_i, A_i)_{i \in [n]}) p(\theta \mid \psi) \stackrel{(iii)}{=} \prod_{i \in [n]}p(R_i \mid \theta, X_i, A_i) p(\theta \mid \psi) \,,  \\
    & \stackrel{(iv)}{=}  \prod_{i \in [n]} \cN(R_i; (u_{A_i} \otimes \phi(X_i))^\top \theta, \sigma^2) \cN(\theta; \W_{\cA} \psi , \Sigma_{\cA})\,,\\
    & = \exp\Big[- \frac{1}{2}
  \Big(v \sum_{i =1}^n (R_i^2 - 2 R_i (u_{A_i} \otimes \phi(X_i))^\top \theta + ((u_{A_i} \otimes \phi(X_i))^\top \theta)^2) + \theta^\top \Lambda_{\cA} \theta - 2 \theta^\top \Lambda_{\cA} \W_{\cA} \psi\\ 
  & \hspace{13cm} + \psi^\top  \W_{\cA} ^\top \Lambda_{\cA} \W_{\cA} \psi\Big)\Big] \,,\\
  &\propto \exp\left[- \frac{1}{2}
  \left( \theta^\top(v\sum_{i =1}^n (u_{A_i} \otimes \phi(X_i)) (u_{A_i} \otimes \phi(X_i))^\top + \Lambda_{\cA}) \theta - 2 \theta^\top \left(v \sum_{i =1}^n (u_{A_i} \otimes \phi(X_i))^\top R_i + \Lambda_{\cA} \W_{\cA} \psi \right)\right)\right]\,,\\
    &= \exp\Big[- \frac{1}{2}
  \Big( \theta^\top(v\sum_{i =1}^n (u_{A_i}u_{A_i}^\top \otimes \phi(X_i) \phi(X_i)^\top)+ \Lambda_{\cA}) \theta - 2 \theta^\top \left(v \sum_{i =1}^n (u_{A_i} \otimes \phi(X_i))^\top R_i + \Lambda_{\cA} \W_{\cA} \psi \right)\Big)\Big]\,,\\
  &\stackrel{(v)}{\propto} \cN\left(\theta; \tilde{\mu}_{\cA}, \left(\tilde{\Lambda}_{\cA}\right)^{-1}\right)\,,
\end{align*}
where we use Bayes rule in $(i)$, $(ii)$ uses two assumptions. First, Given $\theta, X, A$, $R$ is independent of $\psi$. Second, given $\psi$, $\theta$ is independent of $(X, A)$. Moreover, $(iii)$ follows from the assumption that $R_i \mid \theta, X_i, A_i$ are i.i.d. Finally, in $iv$, we replace the distribution by their Gaussian form, and in $(v)$, we set $\tilde{\Lambda}_{\cA} = v\sum_{i =1}^n (u_{A_i}u_{A_i}^\top \otimes \phi(X_i) \phi(X_i)^\top) + \Lambda_{\cA}\,,$ and $  \tilde{\mu}_a = \tilde{\Lambda}_{\cA}^{-1} \big( v \sum_{i =1}^n (u_{A_i} \otimes \phi(X_i))^\top R_i + \Lambda_{\cA} \W_{\cA} \psi\Big)$. Now notice that $\tilde{\Lambda}_{\cA} = {\rm diag}(\Sigma_a^{-1} + G_a)_{a \in \cA}$. Thus, $p(\theta \mid S) = \cN(\theta, \tilde{\mu}_{\cA}, \Lambda_{\cA}^{-1})$ where $\tilde{\mu}_{\cA} = (\tilde{\mu}_{a})_{a \in \cA}$ and $ \Lambda_{\cA} = {\rm diag}(\tilde{\Lambda}_{a})_{a \in \cA}\,,$ with
\begin{align*}
    \tilde{\Lambda}_{a} &= \Sigma_a^{-1} + G_a\,, \nonumber\\
      \tilde{\Lambda}_{a}\tilde{\mu}_{a} &=  \Sigma_a^{-1} \W_{a} \psi + B_a \,.
\end{align*}
The covariance matrix of $p(\theta \mid \psi, S)$ is diagonal by block. Thus $\theta_a \mid \psi, S$ for $a \in \cA$ are independent and have a Gaussian density  $p(\theta_a \mid S) = \cN(\theta_a; \tilde{\mu}_{a}, \tilde{\Sigma}_{a})$ where $\tilde{\Sigma}_{a} = \tilde{\Lambda}_{a}^{-1}$.
\end{proof}

\subsection{Action Posterior}\label{app:marginal-action-posterior}
\begin{proof}[Derivation of 
 $\, p(\theta_a \mid S)$]
We know that $\theta_a \mid S, \psi \sim \cN(\tilde{\mu}_a, \tilde{\Sigma}_a)$ and $\psi \mid S \sim \cN(\bar{\mu}, \bar{\Sigma})$. Thus the posterior density of $\theta_a \mid S$ is also Gaussian since Gaussianity is preserved after marginalization \citep{koller09probabilistic}. We let $\theta_a \mid S \sim \cN(\hat{\mu}_a, \hat{\Sigma}_a)$. Then, we can compute $\hat{\mu}_a$ and $ \hat{\Sigma}_a$ using the total expectation and total covariance decompositions. Let $\Lambda_a =  \Sigma_a^{-1}$. Then we have that
\begin{align*}
    \tilde{\Sigma}_a &= \left( G_a + \Lambda_a \right)^{-1} \\
    \condE{\theta_a}{\psi, S} & = \tilde{\Sigma}_a \left( B_a  + \Lambda_a \W_a \psi \right)
\end{align*}
First, given $S$, $\tilde{\Sigma}_a =\left( G_a + \Lambda_a \right)^{-1}$ and $B_a$ are constant (do not depend on $\psi$). Thus 
\begin{align*}
    \hat{\mu}_a = \condE{\theta_a}{ S}= \E{}{\condE{\theta_a}{\psi, S} \mid S} = \E{\psi \sim \cN(\bar \mu, \bar \Sigma)}{\tilde{\Sigma}_a \left( B_a  + \Lambda_a \W_a \psi \right)} &= \tilde{\Sigma}_a \left( B_a  + \Lambda_a \W_a \E{\psi \sim \cN(\bar \mu, \bar \Sigma)}{\psi} \right)\,,\\
    &= \tilde{\Sigma}_a \left( B_a  + \Lambda_a \W_a \bar \mu \right)\,.
\end{align*}
This concludes the computation of $\hat{\mu}_a$. Similarly, given $S$, $\tilde{\Sigma}_a =\left( G_a + \Lambda_a \right)^{-1}$ and $B_a$ are constant (do not depend on $\psi$), yields two things. First,
\begin{align*}   \condE{\condcov{\theta_a}{\psi, S}}{S} = \condE{\tilde{\Sigma}_a}{S} = \tilde{\Sigma}_a\,.
\end{align*}
Second,
\begin{align*}
    \condcov{\condE{\theta_a}{\psi, S}}{S} &= \condcov{\tilde{\Sigma}_a\Lambda_a \W_a \psi}{S}\\
    &= \tilde{\Sigma}_a \Lambda_a \W_a \condcov{\psi}{S} \W_a^\top \Lambda_a \tilde{\Sigma}_a\\
    &= \tilde{\Sigma}_a \Lambda_a \W_a \bar{\Sigma} \W_a^\top \Lambda_a \tilde{\Sigma}_a\,.
\end{align*}
Finally, the total covariance decomposition \citep{weiss05probability} yields that 
\begin{align*}
 \hat{\Sigma}_a=   \condcov{\theta_a}{S} = \condE{\condcov{\theta_a}{\psi, S}}{S}  +  \condcov{\condE{\theta_a}{\psi, S}}{S} = \tilde{\Sigma}_a+ \tilde{\Sigma}_a \Lambda_a \W_a \bar{\Sigma} \W_a^\top \Lambda_a \tilde{\Sigma}_a\,.
\end{align*}
This concludes the proof.

\end{proof}

\section{MISSING DISCUSSIONS AND PROOFS}\label{sec:ope_proofs}

\subsection{Linear Hierarchy With Non-Linear Rewards}\label{subsec:steps_lin_hierarchy_nonlin_rewards}

In this subsection, the action and latent parameters are generated as in \eqref{eq:contextual_gaussian_model}. But the reward is no longer Gaussian. Precisely, let $ \mathrm{Ber}(p)$ be a Bernoulli distribution with mean $p$, and $g(u) = 1 / (1 + \exp(-u))$ be the logistic function. We define $p(\cdot \mid  x; \theta_a) = \mathrm{Ber}(g(\phi(x)^\top \theta_a))$ which yields
\begin{align}\label{eq:contextual_bernoulli_model}
    \psi & \sim \cN(\mu, \Sigma)\,, \\ 
    \theta_a \mid   \psi& \sim \cN\Big( \W_a \psi , \,  \Sigma_{a}\Big)\,, & \forall a \in \cA\,,\nonumber\\ 
  R \mid \psi, \theta, X, A &\sim \mathrm{Ber}(g(\phi(X)^\top \theta_A))\,.\nonumber
\end{align}

\textbf{Step (A).} In this case, we cannot derive closed-form posteriors. But we can approximate the likelihoods $\LL_a(\cdot)$ by multivariate Gaussian distributions such as 
\begin{align}\label{eq:laplace_approximation}
\LL_a(\cdot) \approx \cN(\cdot; \breve{\mu}_{a}, (\breve{G}_{a})^{-1})\,.
\end{align}
where $\breve{\mu}_{a} $ and $\breve{G}_{a}$ are the maximum likelihood estimator (MLE) and the Hessian of the negative log-likelihood $- \log\LL_a(\cdot)$, respectively:
\begin{align}
        \breve{\mu}_{a} &= \argmax_{\theta \in \real^d} \log \LL_{a}(\theta)\,,\nonumber\\
        \breve{G}_{a} &= \sum_{i \in [n]}\mathds{1}{\{A_i=a\}} \dot{g}\left(X_i^\top \breve{\mu}_{a}\right) \phi(X_i) \phi(X_i)^\top\,.\nonumber
\end{align} 
The latter approximation is slightly different from the popular Laplace approximation\footnote{Laplace approximation's mean is given by the maximum a-posteriori estimate (MAP) of the posterior  and its covariance is the
inverse Hessian of the negative log posterior density.} and is sometimes referred to as Bernstein-von-Mises approximation, see \citep[Remark 3]{schillings2020convergence} for a detailed discussion. It is simpler and more computationally friendly than the Laplace approximation in our setting. Indeed, interestingly, \eqref{eq:laplace_approximation} allows us to use the Gaussian posteriors \eqref{eq:conditional_posterior}, \eqref{eq:effect_posterior} and \eqref{eq:action_posterior_gaussian} in \textbf{Step (A)} in \cref{sec:application}, except that $G_{a}$ and $B_{a}$ are replaced by  $\breve{G}_{a}$ and $\breve{G}_{a}\breve{\mu}_{a}$, respectively (notice that the inverse $\breve{G}_{a}^{-1}$ may not exist but it is not used in our formulas). There is a clear intuition behind this. First, $G_a \gets \breve{G}_{a}$ captures the change of curvature due to the nonlinearity of the mean function $g$. Moreover, $B_{a} \gets \breve{G}_{a}\breve{\mu}_{a}$ follows from the fact that the MLE of the action parameter $\theta_a$ with linear rewards in \eqref{eq:contextual_gaussian_model} is $G_{a}^{-1}B_{a}$, and it corresponds to $\breve{\mu}_{a}$ in this case.

\textbf{Step (B).} The reward function under \eqref{eq:contextual_bernoulli_model} is $r(x, a; \theta) = g(\phi(x)^\top \theta)$. Thus our reward estimate is 
\begin{align*}
    \hat{r}(x, a) =g\Big( \frac{\phi(x)^\top \hat{\mu}_a}{\sqrt{1+\pi/8\normw{\phi(x)}{\hat{\Sigma}_a}}}\Big)\,,
\end{align*}
where we use an approximation of the expectation of the sigmoid function under a Gaussian distribution \citep[Appendix A]{spiegelhalter1990sequential}. 

\textbf{Step (C).} Since $g$ is increasing, the learned policy is \begin{align*}
 \hat{\pi}_{\textsc{g}}(a \mid x) = \mathds{1}{\Big\{a = \argmax_{b \in \cA}  \frac{\phi(x)^\top \hat{\mu}_b}{\sqrt{1+\pi/8\normw{\phi(x)}{\hat{\Sigma}_b}}} \Big\}}.
\end{align*}
\subsection{Frequentist vs. Bayesian Suboptimality}\label{subsec:frequentist_vs_bayes}

Before diving into our main result, a Bayesian suboptimality bound for the structured prior in \eqref{eq:model}, we motivate its use by analyzing both Bayesian and frequentist suboptimality for the simpler non-structured prior in \eqref{eq:basic_model}. While this analysis uses the non-structured prior, our main result applies directly to the structured one. The key distinction between Bayesian and frequentist suboptimality lies in their confidence intervals, as discussed in \citep[Section 4.2]{hong2023multi}. However, we still provide a self-contained explanation for completeness, considering our slightly different results and proofs. Also, note that \citet{hong2023multi} also explored a notion of Bayesian suboptimality that they defined as $V(\pi_*;\theta_*) - V(\hat{\pi};\theta_*) \mid S$. Precisely, they fix the sample set $S$ and consider only randomness in $\theta_*$ sampled from the prior.  In contrast, our definition aligns with traditional online bandits (Bayesian regret) by averaging suboptimality across both the sample set and $\theta_*$ sampled from the prior: $\E{}{V(\pi_*;\theta_*) - V(\hat{\pi};\theta_*)}$. These metrics are different and have distinct implications. Interestingly, ours favors the greedy policy (\cref{subsec:greedy_pessimism}), while theirs incentivizes pessimistic policies.

\textbf{Frequentist confidence intervals.} Assume that for any $a \in \cA$, there exists a parameter $\theta_{*, a} \in \real^d$ generating the reward as $R \mid \theta_*, X, A \sim  \cN(\phi(X)^\top \theta_{*, A}, \sigma^2)$. Then the posterior under the non-structured prior in \eqref{eq:basic_model} satisfies
\begin{equation}\label{eq:frequentist_bound} 
\mathbb{P}\left(\forall x \in \mathcal{X} \,: \,\, |r(x, a; \theta_*)-\hat{r}(x, a)| \leq \left(\alpha(d, \delta) + \frac{c_a + \|\mu_a\|_2}{\sqrt{\lambda_d(\Sigma_a)}}\right) \|\phi(X)\|_{\hat{\Sigma}_a} \mid \tilde{S}\right) \geq 1 - \delta \,,
\end{equation}
where $\tilde{S} = (X_i, A_i)_{i \in [n]}$, the randomness in $ \mathbb{P}$ is over the true reward noise $\cN(0, \sigma^2)$, and $\alpha(d, \delta) = \sqrt{d + 2 \sqrt{d \log \frac{1}{\delta}} + 2 \log \frac{1}{\delta}}$. Now let us compare this confidence interval to the Bayesian one below.

\textbf{Bayesian confidence intervals.} Here we further assume that for any $a \in \cA$, the action parameter $\theta_{*, a}$ is generated from a know Gaussian distribution as $ \theta_{*, a} \sim \cN(\mu_{a}, \Sigma_{a})$ whose parameters $\mu_{a}, \Sigma_{a}$ match the ones used in the non-structured prior \eqref{eq:basic_model}. Then the posterior under the non-structured prior in \eqref{eq:basic_model} satisfies
\begin{align}\label{eq:bayes_bound}
    \condprob{\forall x \in \mathcal{X} \,: \,\, |r(x, a; \theta_*)-\hat{r}(x, a)| \leq \alpha(d, \delta) \|\phi(X)\|_{\hat{\Sigma}_a}}{S} \geq 1-\delta
\end{align}
Here the randomness in $\mathbb{P}$ is over the action parameter uncertainty $\cN(0, \Sigma_{a})$ as we condition on $S = (X_i, A_i, R_i)_{i \in [n]}$. The proof is given after \cref{lemma:bayes_concentration}. In Bayesian bounds, $\theta_{*, a}$  
is treated as a random variable with distribution $\cN(\mu_{a}, \Sigma_{a})$ and it is included in the randomness in $\mathbb{P}$. The randomization of the data $(X_i, A_i, R_i)_{i \in [n]}$ is also included in $\mathbb{P}$. However, we condition on $S$, and hence they are treated as fixed. In contrast with frequentist bounds where 
$\theta_{*, a}$
is fixed while the reward is random and included in the randomness in $\mathbb{P}$ without conditioning on it. The Bayesian confidence intervals capture that the agent has access to the prior on the true action parameters and hence the additional misspecification term $\frac{c_a + \|\mu_a\|_2}{\sqrt{\lambda_d(\Sigma_a)}}$ is not present, as opposed to its frequentist counterpart. This is the main reason why BSO captures the benefits of knowing and leveraging informative priors.

\begin{lemma}[Bayesian bound]\label{lemma:bayes_concentration}
Let $a \in \cA$, $\delta \in (0, 1)$ and $\alpha(d, \delta) = \sqrt{d + 2 \sqrt{d \log \frac{1}{\delta}} + 2 \log \frac{1}{\delta}}$. Then 
\begin{align}
    \condprob{\forall x \in \mathcal{X} \,: \,\, |r(x, a; \theta_*)-r(x, a; \hat{\mu})| \leq \alpha(d, \delta) \|\phi(X)\|_{\hat{\Sigma}_a}}{S} \geq 1-\delta
\end{align}
\end{lemma}

\begin{proof}
First, we have that
\begin{align*}
|r(x, a; \theta_*)-r(x, a; \hat{\mu})|&= |\phi(X)^\top \theta_{*, a}-\phi(X)^\top \hat{\mu}_a|=|\phi(X)^\top\left(\theta_{*, a}-\hat{\mu}_a\right)|\,,\\
&=|\phi(X)^\top \hat{\Sigma}_a^{\frac{1}{2}} \hat{\Sigma}_a^{-\frac{1}{2}}\left(\theta_a-\hat{\mu}_a\right)| \leq\|\phi(X)\|_{\hat{\Sigma}_a}\left\|\theta_a-\hat{\mu}_a\right\|_{\hat{\Sigma}_a^{-1}}\,,
\end{align*}
where we use the Cauchy-Schwarz inequality in the last step. Now we assumed that the true action parameters $\theta_{*, a}$ are random and their prior distribution matches our model in \eqref{eq:contextual_gaussian_model}. Therefore, $\theta_{*, a} \mid S$ have the same density as our posterior $\theta_a \mid S$, and hence $\theta_{*, a} \mid S \sim \cN(\hat{\mu}_a, \hat{\Sigma}_a)$. This means that $\theta_{*, a} - \hat{\mu}_a \mid S \sim \cN(0, \hat{\Sigma}_a)$. Thus, $\hat{\Sigma}_a^{-\frac{1}{2}}(\theta_{*, a} - \hat{\mu}_a) \sim \cN(0, I_d)$. But notice that $\left\|\theta_{*, a}-\hat{\mu}_a\right\|_{\hat{\Sigma}_a^{-1}} = \|\hat{\Sigma}_a^{-\frac{1}{2}}(\theta_{*, a} - \hat{\mu}_a)\|$. Thus we apply \citet[Lemma~1]{laurent2000adaptive} and get that 
\begin{align*}
        \condprob{\left\|\theta_{*, a}-\hat{\mu}_a\right\|_{\hat{\Sigma}_a^{-1}} \leq \alpha(d, \delta)}{S} \geq 1-\delta\,.
\end{align*}
Finally using that for any $x \in \cA$, $|r(x, a; \theta_*)-r(x, a; \hat{\mu})| \leq \|\phi(X)\|_{\hat{\Sigma}_a}\left\|\theta_a-\hat{\mu}_a\right\|_{\hat{\Sigma}_a^{-1}}$ concludes the proof. 
\end{proof}

\begin{lemma}[Frequentist bound]\label{lemma:frequentist_concentration}
Let $a \in \cA$, $\delta \in (0, 1)$ and $\alpha(d, \delta) = \sqrt{d + 2 \sqrt{d \log \frac{1}{\delta}} + 2 \log \frac{1}{\delta}}$. Then 
\begin{align}
    \condprob{\forall x \in \mathcal{X} \,: \,\, |r(x, a; \theta_*)-r(x, a; \hat{\mu})| \leq \left(\alpha(d, \delta) + \frac{c_a + \|\hat{\mu}_a\|_2}{\sqrt{\lambda_d(\Sigma_a)}}\right) \|\phi(X)\|_{\hat{\Sigma}_a}}{\tilde{S}} \geq 1-\delta\,,
\end{align}
where $\tilde{S} = (X_i, A_i)_{i \in [n]}$.
\end{lemma}

\begin{proof}
Similarly to \cref{lemma:bayes_concentration}, we have that $|r(x, a; \theta_*)-r(x, a; \hat{\mu})| \leq \|\phi(X)\|_{\hat{\Sigma}_a}\left\|\theta_a-\hat{\mu}_a\right\|_{\hat{\Sigma}_a^{-1}}$. Note that $(X_i, A_i)$ and $\theta_{*, a}$ are fixed; the randomness only comes from $R_i$ for $i \in [n]$. Thus $\hat{\Sigma}_a$ is fixed while $\hat{\mu}_a$ is random. Keeping this in mind, we have that 
$R_i \sim \cN(\phi(X_i)^\top \theta_{*, a}, \sigma^2)$ and thus 
\begin{align}
    \sum_{i \in [n]}\mathbb{I}_{\{A_i=a\}} \phi(X_i) R_i \sim  \cN\Big(\sum_{i \in [n]}\mathbb{I}_{\{A_i=a\}}\phi(X_i)\phi(X_i)^\top \theta_{*, a}, \sigma^2\sum_{i \in [n]}\mathbb{I}_{\{A_i=a\}}\phi(X_i)\phi(X_i)^\top\Big)\,.
\end{align}
But $ \sum_{i \in [n]}\mathbb{I}_{\{A_i=a\}} \phi(X_i) R_i=\sigma^{2} B_a$ and $\sum_{i \in [n]}\mathbb{I}_{\{A_i=a\}}\phi(X_i)\phi(X_i)^\top=\sigma^{2}G_a$. Thus, we get that 
\begin{align}
  \sigma^{2} B_a \sim  \cN( \sigma^{2} G_a  \theta_{*, a}, \sigma^4 G_a )\,,
\end{align}
and hence
\begin{align}
  B_a \sim  \cN( G_a  \theta_{*, a}, G_a )\,.
\end{align}
Now since $\Sigma_a$, $\mu_a$ and $\hat{\Sigma}_a$ are fixed, we have that 
\begin{align}\label{eq:dist_mu_a_1}
\hat{\mu}_a = \hat{\Sigma}_a(\Sigma_a^{-1} \mu_a + B_a)  \sim  \cN(\hat{\Sigma}_a(\Sigma_a^{-1} \mu_a + G_a  \theta_{*, a}), \hat{\Sigma}_a G_a \hat{\Sigma}_a )\,.
\end{align}
But notice that $G_a  \theta_{*, a} = (G_a + \Sigma_a^{-1})  \theta_{*, a} - \Sigma_a^{-1} \theta_{*, a} = \hat{\Sigma}_a^{-1}\theta_{*, a} - \Sigma_a^{-1} \theta_{*, a}$ and hence \eqref{eq:dist_mu_a_1} becomes
\begin{align}
\hat{\mu}_a \sim  \cN(\hat{\Sigma}_a\Sigma_a^{-1}( \mu_a -  \theta_{*, a}) + \theta_{*, a} , \hat{\Sigma}_a G_a \hat{\Sigma}_a )\,.
\end{align}
Thus 
\begin{align}\label{eq:dist_mu_a_2}
\hat{\mu}_a - \theta_{*, a} \sim  \cN(\hat{\Sigma}_a\Sigma_a^{-1}( \mu_a -  \theta_{*, a}), \hat{\Sigma}_a G_a \hat{\Sigma}_a )\,.
\end{align}
Multiplying by $\hat{\Sigma}_a^{-\frac{1}{2}}$ leads to 
\begin{align}\label{eq:dist_mu_a_3}
\hat{\Sigma}_a^{-\frac{1}{2}}(\hat{\mu}_a - \theta_{*, a}) \sim  \cN(\hat{\Sigma}_a^{\frac{1}{2}}\Sigma_a^{-1}( \mu_a -  \theta_{*, a}), \hat{\Sigma}_a^{\frac{1}{2}} G_a \hat{\Sigma}_a^{\frac{1}{2}} )\,.
\end{align}
Now we rewrite \eqref{eq:dist_mu_a_3} as 
\begin{align}\label{eq:dist_mu_a_4}
\hat{\Sigma}_a^{-\frac{1}{2}}(\hat{\mu}_a - \theta_{*, a}) = \hat{\Sigma}_a^{\frac{1}{2}}\Sigma_a^{-1}( \mu_a -  \theta_{*, a}) + \hat{\Sigma}_a^{\frac{1}{2}} G_a^{\frac{1}{2}} Z\,, \qquad Z \sim \cN(0, I_d)\,.
\end{align}
Thus we have that 
\begin{align}\label{eq:dist_mu_a_5}
\normw{\hat{\Sigma}_a^{-\frac{1}{2}}(\hat{\mu}_a - \theta_{*, a})}{2} &= \normw{\hat{\mu}_a - \theta_{*, a}}{\hat{\Sigma}_a^{-1}} =  \normw{\hat{\Sigma}_a^{\frac{1}{2}}\Sigma_a^{-1}( \mu_a -  \theta_{*, a}) + \hat{\Sigma}_a^{\frac{1}{2}} G_a^{\frac{1}{2}} Z}{2}\,,\nonumber\\
&\leq  \normw{\hat{\Sigma}_a^{\frac{1}{2}}\Sigma_a^{-1}( \mu_a -  \theta_{*, a})}{2} + \normw{\hat{\Sigma}_a^{\frac{1}{2}} G_a^{\frac{1}{2}} Z}{2}\,,\nonumber\\
&\leq \frac{\normw{ \mu_a -  \theta_{*, a}}{2}}{\sqrt{\lambda_d(\Sigma_a)}}  + \normw{Z}{2}\,. 
\end{align}
Finally, since $Z \sim \cN(0, I_d)$, we apply \citet[Lemma~1]{laurent2000adaptive} and get that 
\begin{align*}
        \mathbb{P}(\normw{Z}{2} \leq \alpha(d, \delta)) \geq 1-\delta\,.
\end{align*}
Combining this with \eqref{eq:dist_mu_a_5} leads to  
\begin{align*}
        \condprob{\left\|\theta_{*, a}-\hat{\mu}_a\right\|_{\hat{\Sigma}_a^{-1}} \leq  \frac{\normw{ \mu_a -  \theta_{*, a}}{2}}{\sqrt{\lambda_d(\Sigma_a)}} + \alpha(d, \delta)}{(X_i, A_i)_{i \in [n]}} \geq 1-\delta\,.
\end{align*}
Finally using that for any $x \in \cA$, $|r(x, a; \theta_*)-r(x, a; \hat{\mu})| \leq \|\phi(X)\|_{\hat{\Sigma}_a}\left\|\theta_a-\hat{\mu}_a\right\|_{\hat{\Sigma}_a^{-1}}$ and using that $\|\theta_{*, a}-\hat{\mu}_a\|_2 \leq c_a + \|\hat{\mu}_a\|_2$  concludes the proof. 
\end{proof}

\subsection{Main Result}\label{subsec:main_result_app}
In this section, we prove \cref{thm:main_thm_1}. Recall that we make the following well-specified prior assumption.
\begin{assumption}[Well-specified priors]\label{assum:well-specified}
Action parameters $\theta_{*, a}$ and rewards are drawn from \eqref{eq:contextual_gaussian_model}. 
\end{assumption}
First, given $x \in \cX$, by definition of the optimal policy, we know that it is deterministic. That is, there exists $a_{x, \theta_*} \in [K]$ such that $\pi_*(a_{x, \theta_*} \mid x) = 1$. To simplify the notation and since $\pi_*$ is deterministic, we let $\pi_*(x) =a_{x, \theta_*} $. Also, we know that the greedy policy is deterministic in $\hat{a}_{x} = \argmax_{b \in \cA} \hat{r}(x, b)$. That is $\hat{\pi}_{\textsc{g}}(\hat{a}_{x} \mid x)=1$. Similarly, we let $\hat{\pi}_{\textsc{g}}(x) = \hat{a}_{x}$. Moreover, we let $\Phi(x, a) = e_a \otimes \phi(X) \in \real^{dK}$ where $e_a \in \real^K$ is the indicator vector of action $a$, such that $e_{a, b}=0$ for any $b \in \cA/\{a\}$ and $e_{a, a}=1$. Also, recall that $\hat{\mu}=(\hat{\mu}_a)_{a\in \cA}$ is the concatenation of the posterior means. 
\begin{align*}
      \textsc{Bso}(\hat{\pi}_{\textsc{g}})&=   \E{}{V(\pi_*;\theta_*) - V(\hat{\pi};\theta_*)}\,,\\
      &= \E{}{r(X,  \pi_*(X); \theta_*) - r(X,  \hat{\pi}_{\textsc{g}}(X); \theta_*)  }\,,\\
      &= \E{}{r(X,  \pi_*(X); \theta_*) - r(X,  \hat{\pi}_{\textsc{g}}(X); \hat{\mu}) + r(X,  \hat{\pi}_{\textsc{g}}(X); \hat{\mu}) - r(X,  \hat{\pi}_{\textsc{g}}(X); \theta_*)  }\,,\\
      &\leq \E{}{r(X,  \pi_*(X); \theta_*) - r(X,   \pi_*(X); \hat{\mu}) + r(X,  \hat{\pi}_{\textsc{g}}(X); \hat{\mu}) - r(X,  \hat{\pi}_{\textsc{g}}(X); \theta_*)  }\,,\\
        &\leq \E{}{r(X,  \pi_*(X); \theta_*) - r(X,   \pi_*(X); \hat{\mu})} + \E{}{ r(X,  \hat{\pi}_{\textsc{g}}(X); \hat{\mu}) - r(X,  \hat{\pi}_{\textsc{g}}(X); \theta_*)}\,.
\end{align*}
Now we start by proving that $\E{}{ r(X,  \hat{\pi}_{\textsc{g}}(X); \hat{\mu}) - r(X,  \hat{\pi}_{\textsc{g}}(X); \theta_*)}=0$. This is achieved as follows
\begin{align*}
   \E{}{ r(X,  \hat{\pi}_{\textsc{g}}(X); \hat{\mu}) - r(X,  \hat{\pi}_{\textsc{g}}(X); \theta_*)} &= \E{}{\E{}{ r(X,  \hat{\pi}_{\textsc{g}}(X); \hat{\mu}) - r(X,  \hat{\pi}_{\textsc{g}}(X); \theta_*) \mid X, S}} \,,\\
   &= \E{}{\E{}{ \phi(X)^\top \hat{\mu}_{\hat{\pi}_{\textsc{g}}(X)} - \phi(X)^\top \theta_{*, \hat{\pi}_{\textsc{g}}(X)} \mid X, S}} \,,\\
    &\stackrel{(i)}{=}  \E{}{\E{}{ \Phi(X, \hat{\pi}_{\textsc{g}}(X))^\top \hat{\mu} - \Phi(X, \hat{\pi}_{\textsc{g}}(X))^\top \theta_{*} \mid X, S}} \,,\\
    &\stackrel{(ii)}{=}  \E{}{\Phi(X, \hat{\pi}_{\textsc{g}}(X))^\top \E{}{ \hat{\mu} - \theta_{*} \mid X, S}} \,,\\
    &\stackrel{(iii)}{=}  \E{}{\Phi(X, \hat{\pi}_{\textsc{g}}(X))^\top ( \hat{\mu} - \E{}{\theta_{*} \mid X, S})} \,,\\
    &\stackrel{(iv)}{=}  0 \,.
\end{align*}
In $(i)$, we used that by definition of $\Phi(x, a) = e_a \otimes \phi(X) \in \real^{dK}$, we have $\phi(x)^\top \hat{\mu}_a = \Phi(x, a)^\top \hat{\mu}$ for any $(x, a)$, and the same holds for $\theta_*$. In $(ii)$, we used that $\Phi(X, \hat{\pi}_{\textsc{g}}(X))$ is deterministic given $X$ and $S$. In $(iii)$, we used that $\hat{\mu}$ is deterministic given $X$ and $S$. Finally, in $(iv)$, we used that $\E{}{\theta_{*} \mid X, S} = \E{}{\theta_{*} \mid S} = \hat{\mu}$, which follows from the assumption that $\theta_*$ does not depend on $X$ and the assumption that $\theta_*$ is drawn from the prior, and hence when conditioned on $S$, it is drawn from the posterior whose mean is $\hat{\mu}$. Therefore, $\E{}{ r(X,  \hat{\pi}_{\textsc{g}}(X); \hat{\mu}) - r(X,  \hat{\pi}_{\textsc{g}}(X); \theta_*)}=0$ which leads to 
\begin{align*}
      \textsc{Bso}(\hat{\pi}_{\textsc{g}})& \leq \E{}{r(X,  \pi_*(X); \theta_*) - r(X,   \pi_*(X); \hat{\mu})}\,.
\end{align*}
Now let $\delta \in (0, 1)$, we define the following high-probability events 
\begin{align*}
  E_{a} & =
  \set{\forall x \in \mathcal{X} \,: \,\, |r(x, a; \theta_*)-r(x, a; \hat{\mu})| \leq \alpha(d, \delta) \|\phi(x)\|_{\hat{\Sigma}_a}}\,, &  & \, \forall a \in \cA\,,
\end{align*}
where $\alpha(d, \delta) = \sqrt{d + 2 \sqrt{d \log \frac{1}{\delta}} + 2 \log \frac{1}{\delta}}$. Then we decompose $\textsc{Bso}(\hat{\pi}_{\textsc{g}})$ as
\begin{align*}
      &\textsc{Bso}(\hat{\pi}_{\textsc{g}}) \leq \E{}{r(X,  \pi_*(X); \theta_*) - r(X,   \pi_*(X); \hat{\mu})}\,,\\
      &\leq \E{}{|r(X,  \pi_*(X); \theta_*) - r(X,   \pi_*(X); \hat{\mu})|}\,,\\
     &\leq \E{}{|r(X,  \pi_*(X); \theta_*) - r(X,   \pi_*(X); \hat{\mu})|\I{ E_{\pi_*(X)}}} + \E{}{|r(X,  \pi_*(X); \theta_*) - r(X,   \pi_*(X); \hat{\mu})|\I{ \bar{E}_{\pi_*(X)}}}\,,\\
     &\leq \alpha(d, \delta)\E{}{ \|\phi(X)\|_{\hat{\Sigma}_{\pi_*(X)}}} + \E{}{|r(X,  \pi_*(X); \theta_*) - r(X,   \pi_*(X); \hat{\mu})|\I{ \bar{E}_{\pi_*(X)}}}\,.
\end{align*}
Now we deal with the term $\E{}{|r(X,  \pi_*(X); \theta_*) - r(X,   \pi_*(X); \hat{\mu})|\I{ \bar{E}_{\pi_*(X)}}}$. Let $Z_a = r(X, a; \theta_*) - r(X,   a; \hat{\mu})$, so that $Z_{\pi_*(X)} = r(X,  \pi_*(X); \theta_*) - r(X,   \pi_*(X); \hat{\mu})$. Then we have that
\begin{align*}
    &\E{}{|r(X,  \pi_*(X); \theta_*) - r(X,   \pi_*(X); \hat{\mu})|\I{ \bar{E}_{\pi_*(X)}}} =  \E{}{|Z_{\pi_*(X)}|\I{|Z_{\pi_*(X)}| > \alpha(d, \delta)\|\phi(X)\|_{\hat{\Sigma}_{\pi_*(X)}}}} \,,\\
    & = \E{}{\E{}{|Z_{\pi_*(X)}|\I{|Z_{\pi_*(X)}| > \alpha(d, \delta)\|\phi(X)\|_{\hat{\Sigma}_{\pi_*(X)}}} \mid X, S}}\,,\\
      & \stackrel{(i)}{\leq}  \E{}{\frac{2}{\|\phi(X)\|_{\hat{\Sigma}_{\pi_*(X)}}\sqrt{2 \pi}}
  \int_{u = \alpha(d, \delta) \|\phi(X)\|_{\hat{\Sigma}_{\pi_*(X)}}}^\infty
  u \exp\left[- \frac{u^2}{2\|\phi(X)\|_{\hat{\Sigma}_{\pi_*(X)}}^2}\right] \dif u}\,, \nonumber \\
  & \stackrel{(ii)}{\leq} \E{}{\|\phi(X)\|_{\hat{\Sigma}_{\pi_*(X)}} \frac{2}{\sqrt{2 \pi}}
  \int_{u = \alpha(d, \delta)}^\infty
  u \exp\left[- \frac{u^2}{2}\right] \dif u}\,,\\
   &\stackrel{(iii)}{\leq} \sqrt{\frac{2}{\pi}} \exp\Big(-\frac{\alpha(d, \delta)^2}{2}\Big) \E{}{ \|\phi(X)\|_{\hat{\Sigma}_{\pi_*(X)}}} \,,
\end{align*}
where $(i)$ follows from the facts that $Z_{\pi_*(X)} \mid X, S \sim \cN(0, \normw{\phi(X)}{\hat{\Sigma}_{\pi_*(X)}})$. In $(ii)$, we use the change of variables $u \gets u/\|\phi(X)\|_{\hat{\Sigma}_{\pi_*(X)}} $. Finally, in $(iii)$, we compute the integral. This leads to the following result
\begin{align}
     \textsc{Bso}(\hat{\pi}_{\textsc{g}}) &\leq \left(\sqrt{\frac{2}{\pi}} \exp\Big(-\frac{\alpha(d, \delta)^2}{2}\Big) + \alpha(d, \delta)\right)\E{}{ \|\phi(X)\|_{\hat{\Sigma}_{\pi_*(X)}}}\,.
\end{align}
But this holds for any $\delta \in (0, 1)$. In particular, it holds for $\delta \to 1$ leading to 
\begin{align}
     \textsc{Bso}(\hat{\pi}_{\textsc{g}}) &\leq \left(\sqrt{\frac{2}{\pi}} \exp\Big(-\frac{d}{2}\Big) +\sqrt{d}\right)\E{}{ \|\phi(X)\|_{\hat{\Sigma}_{\pi_*(X)}}}\,,
\end{align}
since $\lim_{\delta \to 1} \alpha(d, \delta) = \alpha(d, 1) = \sqrt{d}$. But $\sqrt{\frac{2}{\pi}}\exp\Big(-\frac{d}{2}\Big) \leq \sqrt{d}$ which concludes the proof.

\subsection{Proof of \cref{thm:main_thm_2}}\label{app:ope_result}

\begin{proof}
    The proof is similar to that of \cref{thm:main_thm_1}. Precisely, we have that 
    \begin{align*}
    {\textsc{Bmse}}(\hat{r}(x, a)) 
    &= \mathbb{E}\big[\big( \hat{r}(x, a) - r(x, a; \theta_*) \big)^2\big]\,,\\
    &=  \mathbb{E}\big[\mathbb{E}\big[\big( \hat{r}(x, a) - r(x, a; \theta_*) \big)^2\mid S\big]\big]\,,\\
    & = \mathbb{E}\big[\mathbb{E}\big[\big( \phi(x)^\top \hat{\mu}_a - \phi(x)^\top \theta_{*, a}\big)^2\mid S\big]\big] \,,\\
    & = \mathbb{E}\big[\mathbb{E}\big[\big( \phi(x)^\top \theta_{*, a} - \phi(x)^\top \hat{\mu}_a \big)^2\mid S\big]\big] \,.
\end{align*}
But we assumed that the true action parameters $\theta_{*, a}$ are random and their prior distribution matches our model in \eqref{eq:contextual_gaussian_model}. Therefore, $\theta_{*, a} \mid S$ have the same density as our posterior $\theta_a \mid S$, and hence $\theta_{*, a} \mid S \sim \cN(\hat{\mu}_a, \hat{\Sigma}_a)$. This means that $\phi(x)^\top\theta_{*, a}  \mid S \sim \cN(\phi(x)^\top \hat{\mu}_a, \|\phi(x)\|_{\hat{\Sigma}_{a}}^2)$. But $\mathbb{E}\big[\big( \phi(x)^\top \theta_{*, a} - \phi(x)^\top \hat{\mu}_a \big)^2\mid S\big]$ is exactly the variance of $\phi(x)^\top\theta_{*, a}  \mid S$ and thus it is equal to $\|\phi(x)\|_{\hat{\Sigma}_{a}}^2$. Taking the expectation concludes the proof.

\end{proof}


\subsection{Optimality of Greedy 
Policies}\label{subsec:greedy_pessimism}
Here, we show that Greedy policy $\hat{\pi}_{\textsc{g}}$ should be preferred to any other choice of policies when considering the BSO as our performance metric. This is because $\hat{\pi}_{\textsc{g}}$ minimizes the BSO. To see this, note that by definition the Greedy policy $\hat{\pi}_{\textsc{g}}$ is deterministic, that is for any context $x \in \cX$, there exists $\hat{a}_{\textsc{g}}$, such that $\hat{\pi}_{\textsc{g}}(\hat{a}_{\textsc{g}}\mid x) =1$. Thus, for any context $x \in \cX$, we simplify the notation by letting $\hat{\pi}_{\textsc{g}}(x)$ denote the action that has a mass equal to $1$. Then, we have that 
\begin{align}\label{eq:helper_11}
 \E{A \sim \hat{\pi}_{\textsc{g}}(\cdot | x)}{\E{\theta_*}{r(x, A; \theta_*)\mid S}} &= \E{\theta_*}{r(x, \hat{\pi}_{\textsc{g}}(x); \theta_*) \mid S} \geq \E{}{ r(x, a; \theta_*)\mid S} & \forall x, a \in \cX \times \cA\,.
\end{align}
where this follows from the definition of $\hat{r}(x, a) = \E{}{ r(x, a; \theta)\mid S}$, the definition of $\hat{\pi}_{\textsc{g}}$ and the fact that $\theta_*$ is sampled from the prior, which leads to $\E{}{ r(x, a; \theta)\mid S} = \E{}{ r(x, a; \theta_*)\mid S}$. Now \eqref{eq:helper_11} holds for any $x \in \cX$ and $a \in \cA$, and hence it holds in expectation under $X\sim \nu$ and $A \sim \pi(\cdot \mid X)$ for any policy $\pi$. That is, 
\begin{align}\label{eq:helper_12}
 \E{X \sim \nu, A \sim \hat{\pi}_{\textsc{g}}(\cdot | X)}{\E{\theta_*}{r(x, A; \theta_*)\mid S}} \geq  \E{X \sim \nu, A \sim \pi(\cdot | X)}{\E{\theta_*}{r(x, A; \theta_*)\mid S}}\,.
\end{align}
Taking another expectation w.r.t. the sample set $S$ and using Fubini's theorem and the tower rule leads to $\E{}{V(\hat{\pi}_{\textsc{g}}; \theta_*)} \geq \E{}{V(\pi; \theta_*)}$ for any stationary policy $\pi$. Then, subtracting $\E{}{V(\pi_*; \theta_*)}$ from both sides of the previous inequality yields that the BSO is minimized by $\hat{\pi}_{\textsc{g}}$ compared to any stationary policy $\pi$, in particular, compared to the policy $\pi_{\textsc{p}}$ induced by pessimism.

\subsection{Discussing the Main Assumption}\label{subsec:assumptions}
In this section, we discuss our main assumption of the existence of the latent parameter $\psi$. Let us take the Gaussian case, where were assume that 
\begin{align}\label{eq:assumption}
    \psi & \sim \cN(\mu, \Sigma)\,, \\ 
    \theta_a \mid   \psi& \sim \cN\Big( \W_a \psi , \,  \Sigma_{a}\Big)\,, & \forall a \in \cA\,.\nonumber
\end{align}
Now we discuss if \eqref{eq:assumption} is as mild as assuming that the action parameters are jointly sampled as $\theta \sim \cN(\mu_{\textsc{act}}, \Sigma_{\textsc{act}})$, where $\mu_{\textsc{act}} \in \real^{dK}$ and $\Sigma_{\textsc{act}} \in \real^{dK \times dK}$. To do so, assume that $\theta \sim \cN(0, \Sigma_{\textsc{act}})$, where we set $\mu_{\textsc{act}}=0$ to simplify notation. Then we want to prove that there exists $\psi \sim \cN(0, I_{d^\prime})$, with $\mu \in \real^{d^\prime}$ for some $d^\prime \geq 1$ and $\W \in \real^{dK \times d^\prime}$ such that $\theta \mid \psi \sim \cN(\W\psi, \tilde{\Sigma})$, where $\tilde{\Sigma}$ is diagonal by $d \times d$ blocks. First, note that $(\theta, \psi)$ is also Gaussian with a mean equal to 0 and covariance $\Omega ={\begin{pmatrix}\Sigma_{\textsc{act}}&\Omega _{2}\\\Omega _{2}^\top&I_{d^\prime}\end{pmatrix}} \in \real^{(dK+d^\prime) \times (dK+d^\prime)}$. Then it suffices to find $\W$ such that $\theta - \W\psi \perp\mkern-10mu\perp \psi$. That is, $\W$ such that ${\rm cov}(\theta - \W\psi, \psi)= {\rm cov}(\theta, \psi)- \W{\rm cov}(\psi, \psi) = \Omega_2 - \W = 0$, which leads to $\W =  \Omega_2$. Using this conditional independence and that $\W = \Omega_2$, we get that the distribution of $\theta - \W\psi \mid \psi $ is the same as the distribution of $\theta - \W\psi $ and it writes $\cN(0, \Sigma_{\textsc{act}} - \W \W^\top)$. Thus, $\theta \mid \psi \sim \cN(\W\psi, \Sigma_{\textsc{act}} - \W \W^\top)$. But $\Sigma_{\textsc{act}}$ is positive-definite and hence there exists $\lambda>0$ such that $\Sigma_{\textsc{act}} - \lambda I_{dK}$ is symmetric and positive semi-definite. Therefore, $\Sigma_{\textsc{act}} - \lambda I_{dK}$ can be written as $\Sigma_{\textsc{act}} - \lambda I_{dK} = \Beta \Beta^\top$ where $\Beta \in \real^{dK \times d^\prime}$ is such that $d^\prime \geq {\rm rank}(\Sigma_{\textsc{act}} - \lambda I_{dK})$ and setting $\W = \Beta$ concludes the proof.

\subsection{Explicit Bound}\label{subsec:explicit_bound}

\textbf{Assumptions.} Before presenting the explicit bound, we introduce some assumptions to simplify the notation, though the results hold under more general conditions.

\begin{itemize}
    \item \textbf{(A0)} We assume $\Sigma_a = \sigma_0^2 I_d$, $\Sigma = \tau^2 I_{d'}$, $\phi(x) = x$, $\normw{x}{2} \leq 1$, and the matrices $\W_a$ are normalized such that $\lambda_1(\W_a \W_a^\top) = \lambda_d(\W_a \W_a^\top) = 1$.
\end{itemize}
Assumption \textbf{(A0)} can be relaxed; the theory holds for any positive definite matrices $\Sigma_a$ and $\Sigma$, bounded $L_2$ norm contexts $x$, and any matrices $\W_a$. Additionally, we make the following necessary assumptions:
\begin{itemize}
\item \textbf{(A1)} Let $G = \mathbb{E}_{X \sim \nu}[X X^\top]$ with $g = \lambda_d(G)$. We assume that $g > 0$.
\item \textbf{(A2)} The fourth moment of $X$ is bounded, and $X$ satisfies $\sqrt{\mathbb{E}[(v^\top X X^\top v)^2]} \leq \mathrm{h} v^\top G v$ for all $v \in \mathbb{R}^d$. Additionally, $X$ and $X \mid A$ follow the same distribution.
\end{itemize}

\begin{theorem}[Explicit Bound]\label{thm:main_thm_app}
Let $\pi_*(x)$ be the optimal action for context $x$. Under \textbf{(A0)},  \textbf{(A1)} and  \textbf{(A2)}, the BSO of \emph{\alg} under the structured prior \eqref{eq:contextual_gaussian_model} satisfies
\begin{align*} 
    \textsc{Bso}(\hat{\pi}_{\textsc{g}}) \leq 2 \sqrt{\mathbb{E}_{X \sim \nu}\left[\frac{d}{ \sigma^{-2} g \alpha_X+ \sigma^{-2}_0} + \frac{\tau^2 \sigma_0^{-4} d}{ \left(\sigma^{-2} g \alpha_X+ \sigma^{-2}_0\right)^2 } + (\sigma_0^2 + \tau^2)d\exp\Big(- \frac{n \pi_0^2(\pi_*(X)|X)}{2}\Big) \right]+ \frac{2 d (\sigma_0^2 + \tau^2)}{n}}\,,\nonumber
\end{align*}
where $    \alpha_x =  \lfloor \frac{\pi_0(\pi_*(x)|x)}{2} n \rfloor -7 \mathrm{h} \sqrt{\lfloor\frac{\pi_0(\pi_*(x)|x)}{2} n \rfloor (d+2 \ln (n)) }$ for any $x \in \cX$ and $a \in \cA$. 
\end{theorem}
\textbf{Scaling with $n$.} Note that
\begin{align}
    &\alpha_X =  \lfloor \frac{\pi_0(\pi_*(x)|x)}{2} n \rfloor -7 \mathrm{h} \sqrt{\lfloor\frac{\pi_0(\pi_*(x)|x)}{2} n \rfloor (d+2 \ln (n)) }\,, & \forall (x, a) \in \cX \times \cA\,.
\end{align}
In particular, with enough data $n$ large enough, there exists a constant $c>0$ such that
\begin{align*}
    &\alpha_x \geq c \pi_0(\pi_*(x)|x) n \,, & \forall (x, a) \in \cX \times \cA\,.
\end{align*}
Therefore, for $n$ large enough, the bound could be expressed as
\begin{align*} 
    \textsc{Bso}(\hat{\pi}_{\textsc{g}}) &= \mathcal{O}\left(\sqrt{\mathbb{E}_{X \sim \nu}\left[\frac{d}{ \pi_0(\pi_*(X)|X)n+ 1} + \frac{d}{ \left(\pi_0(\pi_*(X)|X)n+ 1\right)^2 } + d \exp\Big(- n \pi_0^2(\pi_*(X)|X)\Big) \right]+ \frac{d}{n}}\right)\,,\nonumber\\
    &= \mathcal{O}\left(\sqrt{\mathbb{E}_{X \sim \nu}\left[\frac{d}{ \pi_0(\pi_*(X)|X)n+ 1}  + d \exp\Big(- n \pi_0^2(\pi_*(X)|X)\Big) \right]+ \frac{d}{n}}\right)\,,
\end{align*}
where we omitted constants and used that $\frac{d}{ \left(\pi_0(\pi_*(X)|X)n+ 1\right)^2 } = \mathcal{O}(\frac{d}{ \pi_0(\pi_*(X)|X)n+ 1 })$ leads to the scaling provided in \cref{thm:main_thm}.

\textbf{Proof idea.} From \cref{thm:main_thm_1}, we establish that
   \[
   \textsc{Bso}(\hat{\pi}_{\textsc{g}}) \leq 2\sqrt{d} \E{}{ \|X\|_{\hat{\Sigma}_{\pi_*(X)}}}\,,
   \]
   where we assume that $\phi(x)=x$. Thus we need to control the scaling of the posterior covariance matrix \(\hat{\Sigma}_{\pi_*(X)}\) with \(n\). First, note that for a fixed context \(x\), the only randomness in \(\hat{\Sigma}_{\pi_*(x)}\) originates from the design matrix \(\hat{G}_{\pi_*(x)}\). Previous works simplify this by assuming the data is well-explored \citep{hong2023multi}, i.e., \(\hat{G}_{a} \succ \gamma n G\) for all \(a \in \mathcal{A}\), where \(G = \E{X \sim \nu}{X X^\top}\). We instead only need to control \(\hat{G}_{\pi_*(x)}\), and instead of assuming it satisfies \(\hat{G}_{\pi_*(x)} \succ \gamma n G\), we will consider the events where this happens and the events where this is violated. We can do this based on when $\hat{G}_{\pi_*(x)}$ is the sum of $n$ out-products. The problem, however, is that in this setting, \(\hat{G}_{\pi_*(x)} \) is the sum of \(N_a\) outer products, where the random variable \(N_a\) of the number of times the optimal action \(\pi_*(X)\) appears in the sample set. To address this, we need to control \(N_a\). Fortunately, we notice that $N_a | X$ follows a Binomial distribution with carefully chosen parameters. Using Hoeffding's inequality, we can bound the tails of this distribution, ensuring that the optimal action appears sufficiently often in the dataset. Once the optimal action counts are bounded, the results from \citet{oliveira2016lower} can be adapted to bound \(\hat{G}_{\pi_*(x)}\). Matrix manipulations combined with eigenvalue inequalities then lead to the desired result.
\begin{proof}
From \cref{thm:main_thm_1}, we have that
    \begin{align}
     \textsc{Bso}(\hat{\pi}_{\textsc{g}}) &\leq 2\sqrt{d} \E{}{ \|\phi(X)\|_{\hat{\Sigma}_{\pi_*(X)}}}\,,\\
       &= 2\sqrt{d} \E{}{ \|X\|_{\hat{\Sigma}_{\pi_*(X)}}}\,,\\
     &\leq 2\sqrt{d\, \E{}{ \|X\|^2_{\hat{\Sigma}_{\pi_*(X)}}}}\,,
\end{align}
where we used that the simplifying assumption that $\phi(x) =x$ in the second equality, and Cauchy-Schwartz in the third inequality. Now recall that to simplify, we also assumed that $\Sigma_a = \sigma_0^2 I_d$ for any $a \in \cA$ and that $\Sigma=\tau^2 I_{d'}$. As a result, we have that:
\begin{align*}
    \hat{\Sigma}_a = \tilde{\Sigma}_{a}  + \sigma_0^{-4}\tilde{\Sigma}_{a}  \W_a \bar{\Sigma} \W_a^\top \tilde{\Sigma}_{a}\,,
\end{align*}
and we also have that $\lambda_1(\tilde{\Sigma}_{a}) \leq \sigma_0^2$ and that  
\begin{align}\label{eq:max_eigenvalue_covariance}
    \lambda_1(\hat{\Sigma}_a) & \leq 
\sigma_0^2 + \tau^2 \,, & \forall a \in \cA\,,
\end{align} 
since $\lambda_1(\W_a \W_a^\top) = 1$. These are obtained using Weyl's inequalities.

Now we introduce some quantities:
\begin{itemize}
    \item $N_{a}:$ number of samples where $A_i = a$.
    \item $\epsilon_{x, a} = \mathbb{P}(A=a \mid x) = \pi_0(a\mid x)$ is the probability of choosing the action $a$ given $x$. 
\end{itemize}

Remark that $n-N_a \mid X \sim \text{Bin}(n, 1-\epsilon_{X, a})$. Therefore, via Hoeffding inequality, we can prove that for any $t>0$ and $a \in \cA$
\begin{align*}
    \mathbb{P}(n-N_a - n (1-\epsilon_{X, a}) > t \mid X) \leq \exp(-2t^2/n)\,,
\end{align*}
which simplifies to
\begin{align*}
    \mathbb{P}(N_a< n\epsilon_{X, a} -t \mid X) \leq \exp(-2t^2/n)\,.
\end{align*}
In particular, for $t= n \epsilon_{X, a}/2$, we have that 
\begin{align*}
    \mathbb{P}(N_a < n \epsilon_{X, a}/2 \mid X) \leq \exp(-n \epsilon_{X, a}^2/2)\,.
\end{align*}
If we let $\gamma_{x, a} = \epsilon_{x, a}/2$, we get that 
\begin{align*}
    \mathbb{P}(N_a < \gamma_{X, a} n \mid X)\leq \exp(-2 n \gamma_{X, a}^2)\,.
\end{align*}
Now, we define $\Omega_{x, a} = \{N_a \geq \gamma_{x, a} n\}$, then we have that 
\begin{align}
    \mathbb{P}(\bar{\Omega}_{X, a} \mid X)\leq \exp(-2n \gamma_{X, a}^2)\,.
\end{align}

In particular, 
\begin{align}\label{eq:sdm_bin_concentration}
    \mathbb{P}(\bar{\Omega}_{X, \pi_*(X)} \mid X)\leq \exp(-2n \gamma_{X, \pi_*(X)}^2)\,.
\end{align}
First, we have that
\begin{align}\label{eq:sdm_first_decomposition}
    \sqrt{\mathbb{E}\big[ \|X\|_{\hat{\Sigma}_{\pi_*(X)}}^2\big]} &= \sqrt{\mathbb{E}\big[\mathbb{E}\big[ \|X\|_{\hat{\Sigma}_{\pi_*(X)}}^2\mid X\big]\big]}\,,\nonumber\\
    &= \sqrt{\mathbb{E}\big[\mathbb{E}\big[ \|X\|_{\hat{\Sigma}_{\pi_*(X)}}^2 \mathds{I}\{\Omega_{X, \pi_*(X)}\}\mid X\big]\big] + \mathbb{E}\big[\mathbb{E}\big[ \|X\|_{\hat{\Sigma}_{\pi_*(X)}}^2 \mathds{I}\{\bar{\Omega}_{X, \pi_*(X)}\}\mid X\big]\big]}\,,\nonumber\\
    &= \sqrt{\mathbb{E}\big[I_1\big]+\mathbb{E}\big[I_2\big]}\,.
\end{align}
where $I_1 = \mathbb{E}\big[ \|X\|_{\hat{\Sigma}_{\pi_*(X)}}^2 \mathds{I}\{\Omega_{X, \pi_*(X)}\}\mid X\big]$ and $I_2 = \mathbb{E}\big[ \|X\|_{\hat{\Sigma}_{\pi_*(X)}}^2 \mathds{I}\{\bar{\Omega}_{X, \pi_*(X)}\}\mid X\big]\big]$.

To bound $I_2$, we use the assumption that $\normw{x}{2} \leq 1$ for any context $x$ and \eqref{eq:max_eigenvalue_covariance} leads to 
\begin{align*}
    &\normw{x}{\hat{\Sigma}_a}^2 \leq \sigma_0^2 + \tau^2\,,  & \forall x, a\,.
\end{align*}
In particular, we have that
\begin{align*}
    \normw{x}{\hat{\Sigma}_{\pi_*(x)}}^2 \leq  \sigma_0^2  + \tau^2\,.
\end{align*}
Let $c_1 =  \sigma_0^2  + \tau^2$. Then we have that 
\begin{align}\label{eq:sdm_helper2}
    I_2 &= \mathbb{E}\big[ \|X\|_{\hat{\Sigma}_{\pi_*(X)}}^2 \mathds{I}\{\bar{\Omega}_{X, \pi_*(X)}\}\mid X\big]\,,\nonumber\\
    &\leq \mathbb{E}\big[ c_1 \mathds{I}\{\bar{\Omega}_{X, \pi_*(X)}\}\mid X\big]\,,\nonumber\\
     &= c_1\mathbb{E}\big[  \mathds{I}\{\bar{\Omega}_{X, \pi_*(X)}\}\mid X\big]\,,\nonumber\\
    &= c_1 \mathbb{P}(\bar{\Omega}_{X, \pi_*(X)} \mid X) \,,\nonumber\\
    &\leq c_1 \exp(-2n \gamma_{X, \pi_*(X)}^2)
\end{align}

\begin{lemma}[\citet{oliveira2016lower}]\label{lemma:covariance_concentration}
Assume $X_1, \ldots, X_n \in \mathbb{R}^{d}$ are i.i.d. random variables with bounded fourth moments. Define $G \equiv \mathbb{E}\left[X_1X_1^\top\right]$ and $\widehat{G}_n \equiv \sum_{i=1}^n X_iX_i^\top$. Let $\mathrm{h} \in(1,+\infty)$ be such that $\sqrt{\mathbb{E}\left[\left(v^T X_1X_1^\top v\right)^2\right]} \leq \mathrm{h} v^T G v$ for all $v \in \mathbb{R}^d$. Then for any $\delta \in(0,1)$:
\begin{align*}
    \mathbb{P}\left( \widehat{G}_n \succeq  \alpha(n, \delta) G \right) \geq 1-\delta\,,
\end{align*}
where $\alpha(n, \delta) =  n-7 \mathrm{h} \sqrt{n (d+2 \ln (2 / \delta))}$. In particular, for any fixed positive definite matrix $\Lambda$, we have that
\begin{align}\label{eq:sdm_lemma_main_eq}
    \mathbb{P}\left( (\widehat{G}_n + \Lambda)^{-1} \preceq  (\alpha(n, \delta)  G+ \Lambda)^{-1}\right)   \geq 1-\delta\,,
\end{align}
\end{lemma}

Let us focus on the firm term $I_1$. Thanks to multiplication with $\mathds{I}\{\Omega_{X, \pi_*(X)}\}$, we know that given $X$, $N_{\pi_*(X)} \geq \gamma_{X, \pi_*(X)}\, n$. Therefore, given $X$, we have that 
\begin{align}
    &\hat{G}_{\pi_*(X)} = \sigma^{-2} \sum_{i \in [n]} \mathbb{I}_{\{A_i=\pi_*(X)\}} X_i X_i^\top \succeq \sigma^{-2} \sum_{i=1}^{\lfloor \gamma_{X, \pi_*(X)}\, n \rfloor} X^\prime_i {X^\prime_i}^\top\,, &\forall a \in \cA\,,
\end{align}
where the $X_i^\prime$ are copies of $X_i$ (assuming that $X|A$ and $X$ have the same law). To ease exposition, let $\alpha(\lfloor \gamma_{x, \pi_*(x)} n \rfloor, \delta) = \alpha_{x}$. Then, applying \eqref{eq:sdm_lemma_main_eq} in \cref{lemma:covariance_concentration} with $\Lambda = \Sigma_0^{-1} = \sigma_0^{-2} I_d$ and $\delta$ leads to 
\begin{align}
    \mathbb{P}\left( (\hat{G}_{\pi_*(X)} + \sigma_0^{-2} I_d)^{-1} \mathds{I}\{\Omega_{X, \pi_*(X)}\}\preceq  (\sigma^{-2}\alpha_X G+ \sigma_0^{-2} I_d)^{-1} \mid X\right)   \geq 1-\delta\,,
\end{align}
where $G = \E{X \sim \nu}{X X^\top}$. Noticing that $\tilde{\Sigma}_{\pi_*(X)} = (\hat{G}_{\pi_*(X)} + \sigma_0^{-2} I_d)^{-1}$ leads to
\begin{align}
    \mathbb{P}\left(\tilde{\Sigma}_{\pi_*(X)} \mathds{I}\{\Omega_{X, \pi_*(X)}\} \leq  (\sigma^{-2}\alpha_X  G+ \sigma_0^{-2}I_d)^{-1} \mid X\right)   \geq 1-\delta\,.
\end{align}
In particular, we bound its maximum eigenvalue as
\begin{align}\label{eq:sdm_main_concentration}
    \mathbb{P}\left( \lambda_1(\tilde{\Sigma}_{\pi_*(X)})\mathds{I}\{\Omega_{X, \pi_*(X)}\} \leq  \frac{1}{ \sigma^{-2} g \alpha_X+ \sigma^{-2}_0} \mid X\right)   \geq 1-\delta\,.
\end{align}
where $g = \lambda_d(G)$.

Moreover, we know that:
\begin{align*}
    \hat{\Sigma}_a = \tilde{\Sigma}_{a}  + \sigma_0^{-4}\tilde{\Sigma}_{a}  \W_a \bar{\Sigma} \W_a^\top \tilde{\Sigma}_{a}\,,
\end{align*}
and thus 
\begin{align*}
    \lambda_1(\hat{\Sigma}_a) = \lambda_1(\tilde{\Sigma}_{a})  + \sigma_0^{-4} \tau^2 \lambda_1(\tilde{\Sigma}_{a})^2  \,,
\end{align*}
where we use Weyl’s inequality and the fact that for any matrix $\Alpha$ and any positive semi-definite matrix $\Beta$ such that the product $\Alpha^\top \Beta \Alpha$ exists, the following inequality holds  $\lambda_1(\Alpha^\top \Beta \Alpha) \leq \lambda_1(\Beta ) \lambda_1(\Alpha^\top \Alpha)$. Therefore, from \eqref{eq:sdm_main_concentration}, we have that 
\begin{align}
    \mathbb{P}\left(\lambda_1(\hat{\Sigma}_{\pi_*(X)})\mathds{I}\{\Omega_{X, \pi_*(X)}\} \leq  \frac{1}{ \sigma^{-2} g \alpha_X+ \sigma^{-2}_0} + \frac{\sigma_0^{-4} \tau^2}{ \left(\sigma^{-2} g \alpha_X+ \sigma^{-2}_0\right)^2} \mid X\right)   \geq 1-\delta\,.
\end{align}
Using the law of total expectation, we get that
\begin{align*}
    I_1 &= \mathbb{E}\big[ \|X\|_{\hat{\Sigma}_{\pi_*(X)}}^2 \mathds{I}\{\Omega_{X, \pi_*(X)}\}\mid X\big]\,,\\
    & \leq \frac{1}{ \sigma^{-2} g \alpha_X+ \sigma^{-2}_0} + \frac{\sigma_0^{-4} \tau^2}{ \left(\sigma^{-2} g \alpha_X+ \sigma^{-2}_0\right)^2} + c_1 \delta \,.
\end{align*}
Setting $\delta = 2/n$ leads to 
\begin{align}\label{eq:sdm_helper1}
    I_1  & \leq \frac{1}{ \sigma^{-2} g \alpha_X+ \sigma^{-2}_0} + \frac{\sigma_0^{-4} \tau^2}{ \left(\sigma^{-2} g \alpha_X+ \sigma^{-2}_0\right)^2} + \frac{2 c_1}{n}\,.
\end{align}

Combining \eqref{eq:sdm_helper2} and \eqref{eq:sdm_helper1} yields
\begin{align}\label{eq:unsimplified}
    \mathbb{E}\big[ \|X\|_{\hat{\Sigma}_{\pi_*(X)}}\big] &\leq \sqrt{\mathbb{E}\big[I_1\big]+\mathbb{E}\big[I_2\big]}\,,\\
    &\leq \sqrt{\mathbb{E}_{X \sim \nu}\left[\frac{1}{ \sigma^{-2} g \alpha_X+ \sigma^{-2}_0}  + \frac{\sigma_0^{-4} \tau^2}{ \left(\sigma^{-2} g \alpha_X+ \sigma^{-2}_0\right)^2} \right] + \frac{2 c_1}{n} +c_1\mathbb{E}_{X \sim \nu}\left[\exp(-2n \gamma_{X, \pi_*(X)}^2)\right]}\,,\nonumber
\end{align}
Replacing $c_1 = \sigma_0^2 + \tau^2$ leads to
\begin{align}\label{eq:simplified}
    &\mathbb{E}\big[ \|X\|_{\hat{\Sigma}_{\pi_*(X)}}\big] \\
    &\leq \sqrt{\mathbb{E}_{X \sim \nu}\left[\frac{1}{ \sigma^{-2} g \alpha_X+ \sigma^{-2}_0}  + \frac{\sigma_0^{-4} \tau^2}{ \left(\sigma^{-2} g \alpha_X+ \sigma^{-2}_0\right)^2} \right] + \frac{2 (\sigma_0^2 + \tau^2)}{n} +(\sigma_0^2 + \tau^2)\mathbb{E}_{X \sim \nu}\left[\exp(-2n \gamma_{X, \pi_*(X)}^2)\right]}\,,\nonumber
\end{align}
Finally, noticing that $\gamma_{X, \pi_*(X)} = \frac{\pi_0(\pi_*(X)|X)}{2}$ concludes the proof.
\end{proof}

\section{ADDITIONAL EXPERIMENTS}\label{app:experiments}
As mentioned in \cref{app:experiments}, our experiments were conducted on internal machines with 30 CPUs and thus they required a moderate amount of computation. These experiments are also reproducible with minimal computational resources.

\subsection{Implementation Details of Baselines}\label{app:details_imp}

We implement the baselines as follows.

\begin{itemize}
    \item \textbf{IPS.} For OPE, we use
    \begin{align}
    &\hat{V}(\pi, S) = \frac{1}{n} \sum_{i=1}^n \frac{\pi(a_i|x_i)}{\max\{\pi_0(a_i|x_i), \tau\}} r_i \,,
\end{align}
and for OPL, we use
    \begin{align}
    &\argmax_\pi \frac{1}{n} \sum_{i=1}^n \frac{\pi(a_i|x_i)}{\max\{\pi_0(a_i|x_i), \tau\}} r_i \,,
\end{align}
where $\tau \in [0, 1]$ is a hyper-parameter.
    \item \textbf{snIPS.} For OPE, we use
        \begin{align}
    &\hat{V}(\pi, S) = \frac{1}{\sum_{i=1}^n \frac{\pi(a_i|x_i)}{\pi_0(a_i|x_i)}} \sum_{i=1}^n \frac{\pi(a_i|x_i)}{\pi_0(a_i|x_i)} r_i \,,
\end{align}
and for OPL, we use
    \begin{align}
    \argmax_\pi \frac{1}{\sum_{i=1}^n \frac{\pi(a_i|x_i)}{\pi_0(a_i|x_i)}} \sum_{i=1}^n \frac{\pi(a_i|x_i)}{\pi_0(a_i|x_i)} r_i\,,
\end{align}

    \item \textbf{MIPS.} We cluster actions into $L$ groups and let $\phi(a)$ be the cluster of action $a$. Let $c_i$ be the cluster of action $a_i$, then For OPE, we use
        \begin{align}
    &\hat{V}(\pi, S) = \frac{1}{n} \sum_{i=1}^n \frac{\pi(c_i|x_i)}{\pi_0(c_i|x_i)}r_i \,,
\end{align}
where $c_i = \phi(a_i)$ for any $ i\in [n]$, and $\pi(c|x) = \sum_{a \in \mathcal{A}}\mathds{1}\left[\phi(a) = c \right]\pi(a|x)$. For OPL, we use
    \begin{align}
    \argmax_\pi \frac{1}{n} \sum_{i=1}^n \frac{\pi(c_i|x_i)}{\pi_0(c_i|x_i)}r_i \,,
\end{align}
    \item \textbf{PC.} We use the Knn implementation of \texttt{PC}. Let $\psi(a, k)$ be the set of k-nearest neighbors of $a$, then For OPE, we use
        \begin{align}
    &\hat{V}(\pi, S) = \frac{1}{n} \sum_{i=1}^n \frac{\sum_{a \in \mathcal{A}}\mathds{1}\left[a \in \psi(a_i, k) \right]\pi(a|x)}{\sum_{a \in \mathcal{A}}\mathds{1}\left[a \in \psi(a_i, k)\right]\pi_0(a|x)}r_i \,,
\end{align}
    \begin{align}
    \argmax_\pi \frac{1}{n} \sum_{i=1}^n \frac{\sum_{a \in \mathcal{A}}\mathds{1}\left[a \in \psi(a_i, k) \right]\pi(a|x)}{\sum_{a \in \mathcal{A}}\mathds{1}\left[a \in \psi(a_i, k)\right]\pi_0(a|x)}r_i \,.
\end{align}

    \item \textbf{DM (Freq).} This DM uses the linear-Gaussian likelihood model $R \mid \theta, X, A \sim \cN(\phi(X)^\top \theta_A , \sigma^2)$ and learn the parameters $\theta_a$ using the maximum likelihood principle leading to 
    \begin{align}
        \hat{r}(x, a) = \phi(X)^\top \hat{\mu}_a\,,
    \end{align}
    where the MLE is \(\hat{\mu}_{a} = (G_a + \lambda I_d)^{-1} B_a \), with \( G_{a} = \sum_{i \in [n]} \mathbb{I}_{\{A_i=a\}} \phi(X_i) \phi(X_i)^\top \) and \( B_{a} = \sum_{i \in [n]} \mathbb{I}_{\{A_i=a\}} R_i \phi(X_i) \), and $\lambda$ is a regularization hyper-parameter.
    
    \item \textbf{DM (Bayes).} This DM uses the linear-Gaussian likelihood model combined with Gaussian priors as
\begin{align}
  \theta_a &\sim \cN(\mu_{a}, \Sigma_{a})\,, & \forall a \in \cA\,,\\
    R \mid \theta, X, A &\sim \cN(\phi(X)^\top \theta_A , \sigma^2)\,,\nonumber
\end{align}
Under this prior, each action \( a \) has an associated parameter \( \theta_a \). Given the prior in \eqref{eq:basic_model}, the posterior distribution of an action parameter follows a multivariate Gaussian: \( \theta_a \mid S \sim \mathcal{N}(\hat{\mu}_a, \hat{\Sigma}_a) \), where \( \hat{\Sigma}_{a}^{-1} = \Sigma_{a}^{-1} + G_{a} \) and \( \hat{\Sigma}_{a}^{-1} \hat{\mu}_{a} = \Sigma_{a}^{-1} \mu_a + B_a \). Here, \( G_{a} = \sigma^{-2} \sum_{i \in [n]} \mathbb{I}_{\{A_i=a\}} \phi(X_i) \phi(X_i)^\top \) and \( B_{a} = \sigma^{-2} \sum_{i \in [n]} \mathbb{I}_{\{A_i=a\}} R_i \phi(X_i) \). Then, the reward estimate is 
    \begin{align}
        \hat{r}(x, a) = \phi(X)^\top \hat{\mu}_a\,,
    \end{align}
    \item \textbf{DR.} For OPE, we use
    \begin{align}
    \hat{V}(\pi, S) = \frac{1}{n} \sum_{i=1}^n \frac{\pi(a_i|x_i)}{\max\{\pi_0(a_i|x_i), \tau\}} \left(r_i - \hat{r}(a_i, x_i) \right)+
 \E{a \sim \pi(\cdot | x_i)}{\hat{r}(x_i, a)} \,,
\end{align}
with $\tau \in [0, 1]$ and $\hat{r}$ is the reward model obtained using \texttt{DM (Freq)}. For OPL, we use
    \begin{align}
    \argmax_\pi \frac{1}{n} \sum_{i=1}^n &\frac{\pi(a_i|x_i)}{\max\{\pi_0(a_i|x_i), \tau\}} \left(r_i - \hat{r}(a_i, x_i) \right)+
 \E{a \sim \pi(\cdot | x_i)}{\hat{r}(x_i, a)}\,,
\end{align}
\end{itemize}

\subsection{Additional Results}\label{app:ope_experiments}
\textbf{Additional Synthetic Experiments.} We provide additional OPE and OPL results. We consider the same synthetic setting as in \cref{sec:experiments} and we vary $K \in \{100, 1000\}$ and $d' \in \{5, 10, 20\}$. 

For OPE, we fix a target policy \(\pi\) and assess algorithm performance in evaluating the value function of that target policy using the mean squared error (MSE) averaged over 50 problem instances sampled from the prior. This can be seen as a proxy for our theoretical metric, Bayesian MSE (BMSE). The target policy is defined as \(\epsilon\)-greedy with \(\epsilon=0.5\); it chooses the best action with probability 0.5 and a random action with probability 0.5.

\begin{figure}[H]
  \centering  \includegraphics[width=\linewidth]{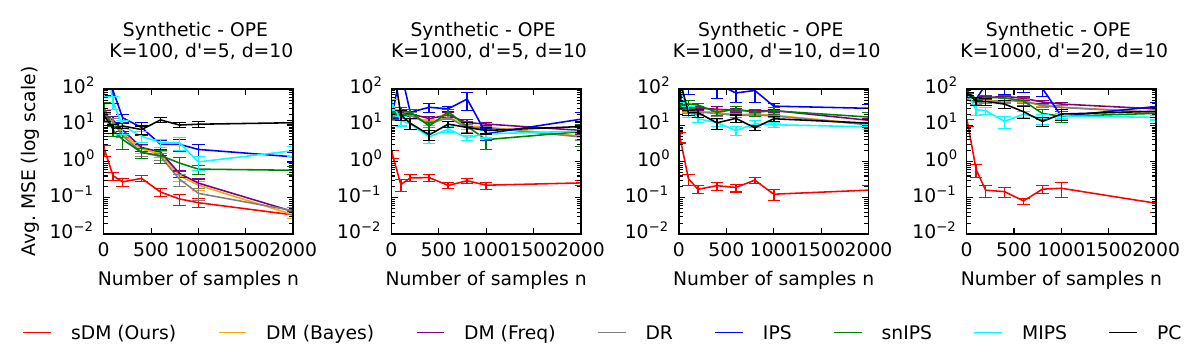}
  \caption{The average MSE of an $\epsilon$-greedy target policy on \textbf{synthetic problems} with varying $n$, $K$ and $d^\prime$.} 
  \label{fig:ope_results}
\end{figure}

Results are shown in \cref{fig:ope_results}. While \alg maintains its advantage over baselines, particularly in low-data settings, an interesting observation emerged. IPS-based methods (\texttt{snIPS}, \texttt{DR} and \texttt{MIPS}) outperformed standard direct methods (\texttt{DM (Bayes)} and \texttt{DM (Freq)}) in the OPE experiments, which was not the case in OPL experiments. The only direct method that outperformed them in OPE is \alg. Also, \texttt{MIPS} and \texttt{PC} outperform other IPS-variants when the number of action is $1000$, but fail to outperform standard methods when $K=100$.

For OPL, the gap in performance between \alg and all other baselines widens when $K$ increases. It also widens when the problem becomes higher-dimensional (e.g., increased value of $d'$).

\begin{figure}[H]
  \centering  \includegraphics[width=\linewidth]{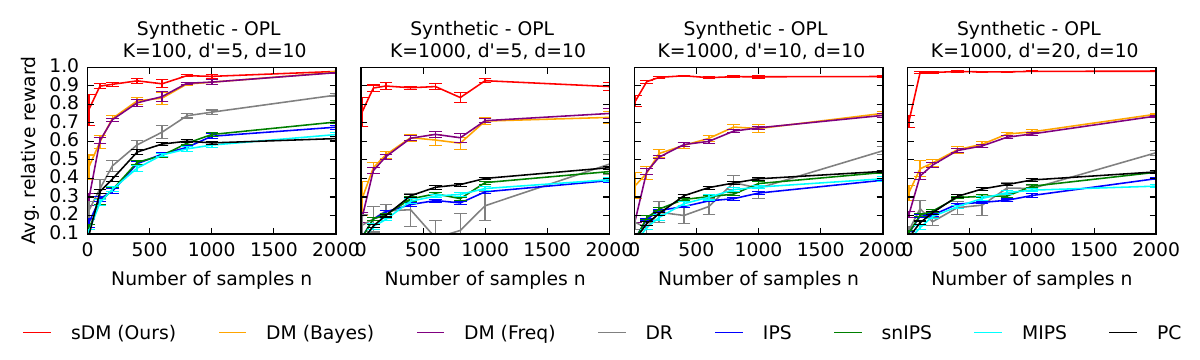}
  \caption{The average relative reward of the learned policy using one of the baselines on \textbf{synthetic problems} with varying $n$, $K$ and $d^\prime$.} 
  \label{fig:opl_results}
\end{figure}

\textbf{Additional MovieLens Experiments.} Similar conclusions are drawn as in the main text.

\begin{figure}[H]
  \centering  \includegraphics[width=0.7\linewidth]{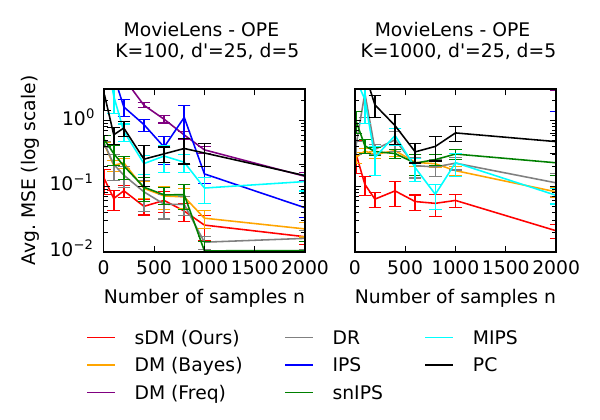}
  \caption{The average MSE of an $\epsilon$-greedy target policy on \textbf{MovieLens problems} with varying $n$, $K$ and $d^\prime$.} 
\end{figure}

\begin{figure}[H]
  \centering  \includegraphics[width=0.7\linewidth]{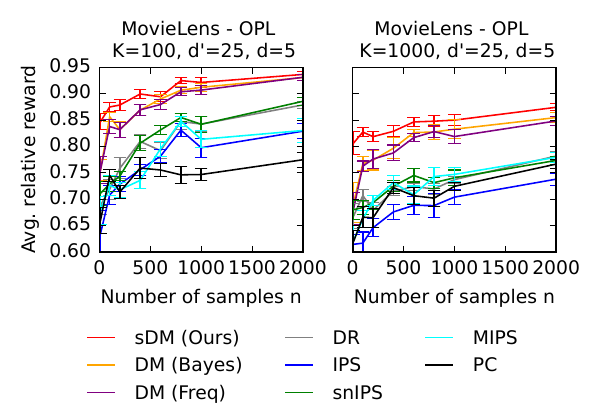}
  \caption{The average relative reward of the learned policy using one of the baselines on \textbf{MovieLens problems} with varying $n$, $K$ and $d^\prime$.} 
\end{figure}

\subsection{Different Logging Policy: $\epsilon$-Greedy}\label{app:epsilon_greedy}
\begin{figure}[H]
  \centering  \includegraphics[width=\linewidth]{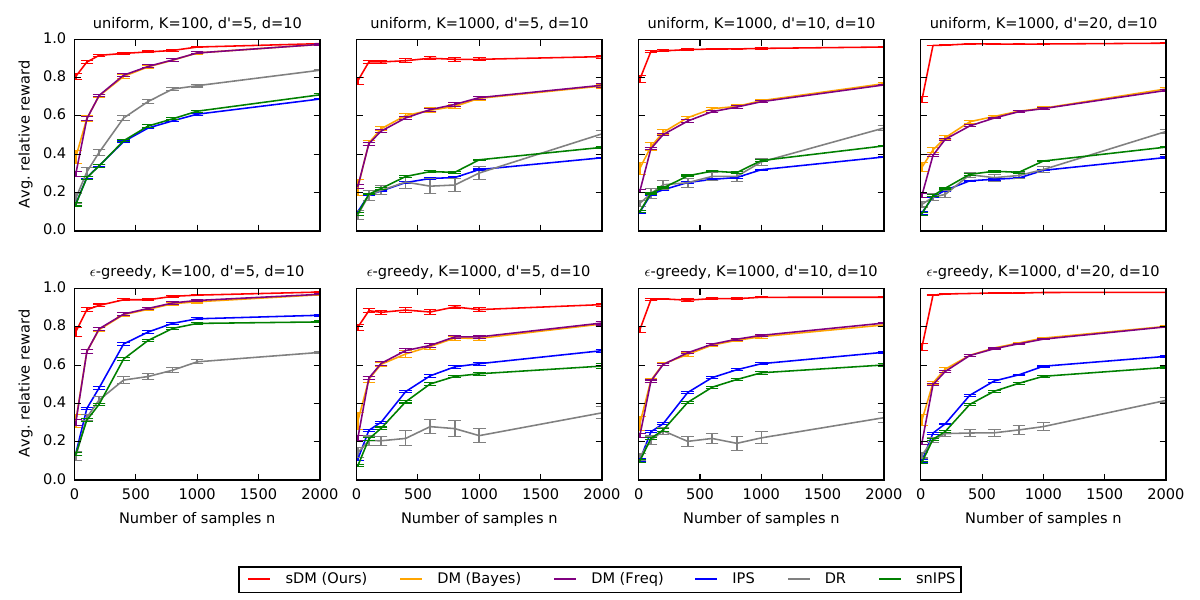}
  \caption{The relative reward of the learned policy on \textbf{synthetic problems} with varying $n$, $K$ and $d^\prime$ and varying logging policies.} 
  \label{fig:synthetic_linear_app}
\end{figure}

\begin{figure}[H]
  \centering  \includegraphics[width=\linewidth]{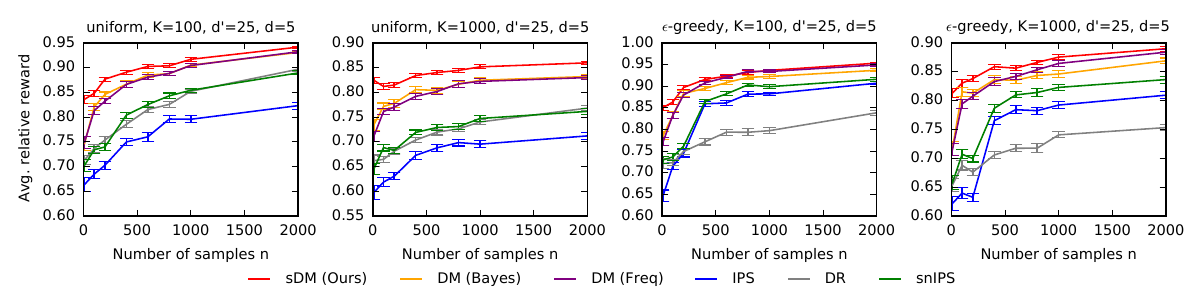}
  \caption{The relative reward of the learned policy on \textbf{MovieLens problems} with varying $n$ and $K$ and varying logging policies.} 
  \label{fig:movielens_linear_app}
\end{figure}

Here we consider the setting in \cref{sec:experiments} and provide additional results with the $\epsilon$-greedy logging policy, where $\epsilon=0.5$. We consider both synthetic and MovieLens datasets. The results are shown in \cref{fig:synthetic_linear_app}. The conclusions are similar to those in \cref{sec:experiments}, except that \texttt{IPS} outperforms the other variants \texttt{DR} and \texttt{snIPS} when the logging policy is performing well. Also, the results for the MovieLens problems are given in \cref{fig:movielens_linear_app} and the conclusions are similar to those made in \cref{sec:experiments}.

\subsection{Robustness to Likelihood Misspecification}\label{app:misspecification_experiments}

We strengthened our evaluation by assessing \alg's robustness to likelihood misspecification below (robustness to prior misspecification is provided in \cref{app:mips_experiments}). In these experiments, the true data-generating process (same as the synthetic experiments in \cref{sec:experiments}) differed from \alg's assumptions in two different ways: either the likelihood is misspecified

\begin{figure}[H]
  \centering  \includegraphics[width=\linewidth]{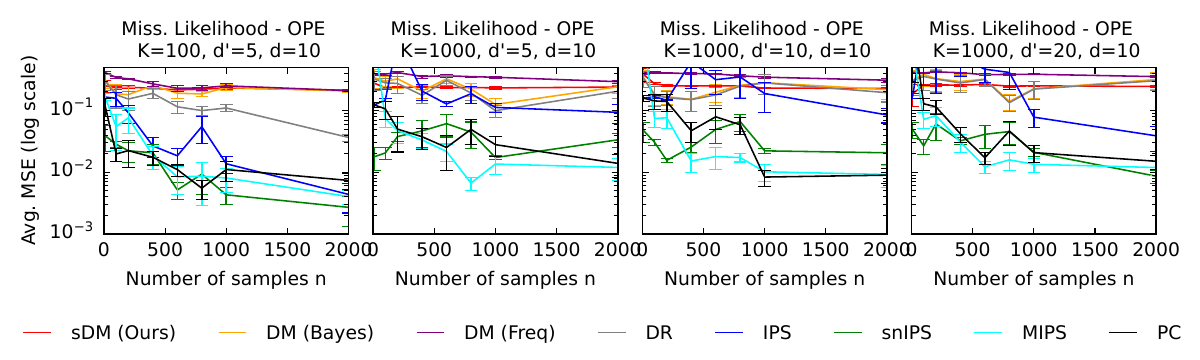}
  \caption{Effect of prior mean and covariance misspecification: The average MSE of an $\epsilon$-greedy target policy on synthetic problems using both misspecified prior means and covariances with varying $n$ and $K$.} 
  \label{fig:likelihood_misspecfication_ope}
\end{figure}

\begin{figure}[H]
\begin{center}
\centerline{\includegraphics[width=\linewidth]{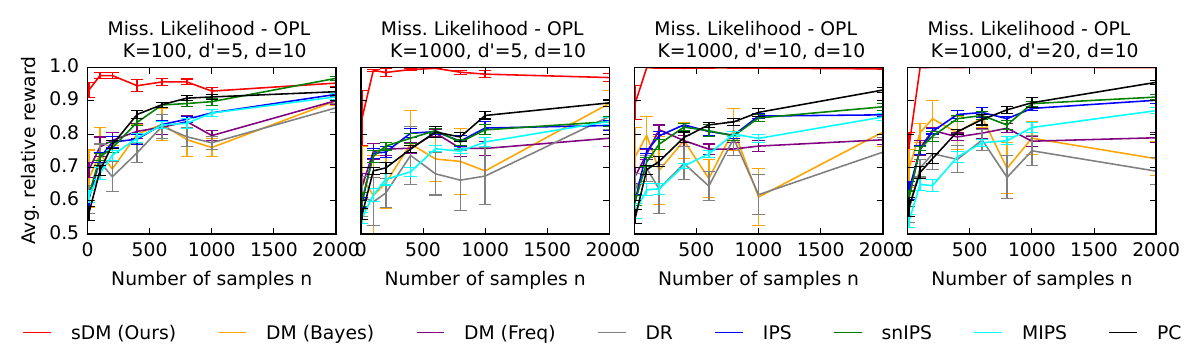}}
\vskip -0.1in
\caption{Effect of likelihood misspecification: The relative reward of the learned policy on synthetic problems using misspecified likelihood with varying $n$ and $K$.}
\label{fig:likelihood_misspecfication_opl}
\end{center}
\end{figure}

\textbf{Misspecified likelihood (\cref{fig:likelihood_misspecfication_ope,fig:likelihood_misspecfication_opl}).} We also simulate when the true reward distribution differed from the likelihood assumed by \alg. For example, we simulated binary rewards using a Bernoulli-logistic model while \alg used a linear-Gaussian likelihood. Other DMs: \texttt{DM (Bayes)} and \texttt{DM (Freq)} also use a misspecified likelihood model and to emphasize this we add the suffix \texttt{Lin} to all DMs names. Overall, we make the same observations as those made in the main text. Likelihood misspecification affect \alg's performance on OPE, but \alg still outperforms all methods by a large margin in OPL.

\subsection{Robustness to Prior Misspecification}\label{app:mips_experiments}

\begin{figure}[H]
  \centering  \includegraphics[width=\linewidth]{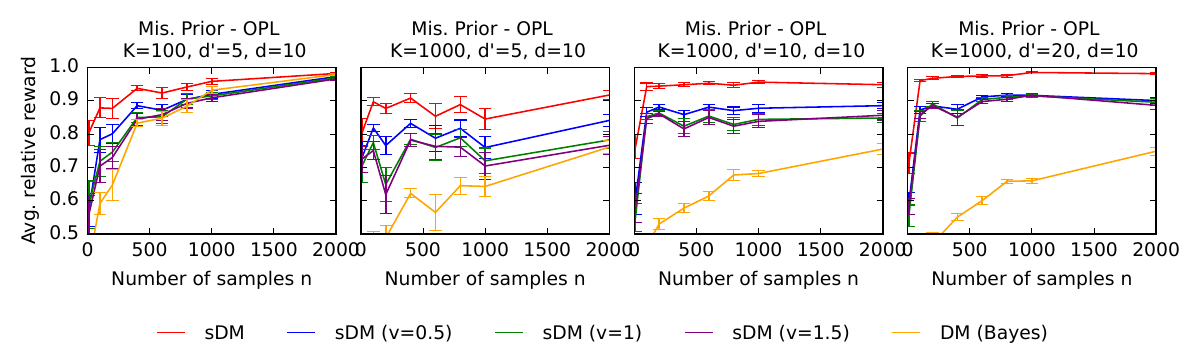}
  \caption{Effect of prior mean and covariance misspecification: The average relative reward of the learned policy on synthetic problems using both misspecified prior means and covariances with varying $n$ and $K$ and $d'$.} 
  \label{fig:mis_prior_mean_covariance}
\end{figure}

\textbf{Misspecified prior means and covariances(\cref{fig:mis_prior_mean_covariance}).}  This is achieved by adding uniformly sampled noise from \([v, v+0.5]\) to both the true prior mean and covariance parameters \(\mu, \Sigma, \W_a, \Sigma_a\), with \(v\) controlling the level of misspecification. We varied \(v \in \{0.5, 1, 1.5\}\) and analyzed its impact on \alg's performance. For comparison, we included the well-specified \alg and the most competitive baseline, \texttt{DM (Bayes)}, while omitting other baselines to reduce clutter. \alg's performance decreases with increasing misspecification, yet \alg with misspecification still outperforms the most competitive baseline, especially when \(K\) is large. We also observe that the impact of prior covariance misspecification is less significant compared to prior mean misspecification.

\subsection{Comparison of Greedy and Pessimistic Policies}\label{app:greedy_pessimism_experiments}

To validate our theory that a greedy policy should be preferred over the commonly adopted pessimistic policy in our Bayesian setting, we used a performance metric averaged over multiple bandit problems sampled from the prior. To verify this, we considered the same OPL synthetic setting as in \cref{sec:experiments} and compared \alg with a greedy policy to \alg with a pessimistic policy. Recall that a greedy policy with respect to our reward estimate writes
\begin{align}
     \hat{\pi}_{\textsc{g}}(a \mid x) = \mathds{1}{\{a = \argmax_{b \in \cA} \hat{r}(x, b)\}}\,,
\end{align}
while a pessimistic one writes
\begin{align}
     \hat{\pi}_{\textsc{p}}(a \mid x) = \mathds{1}{\{a = \argmax_{b \in \cA} \hat{r}(x, b) - u(x, a)\}}\,,
\end{align}
where $u(x, a) = \alpha(d, \delta) \|\phi(X)\|_{\hat{\Sigma}_a}$ with $\alpha(d, \delta) = \sqrt{d + 2 \sqrt{d \log \frac{1}{\delta}} + 2 \log \frac{1}{\delta}}$
and this is derived in \eqref{eq:bayes_bound} in \cref{subsec:frequentist_vs_bayes}. As predicted by our theory, the results show that the greedy policy has better average performance over multiple bandit instances sampled from the prior.

\begin{figure}[H]
  \centering  \includegraphics[width=\linewidth]{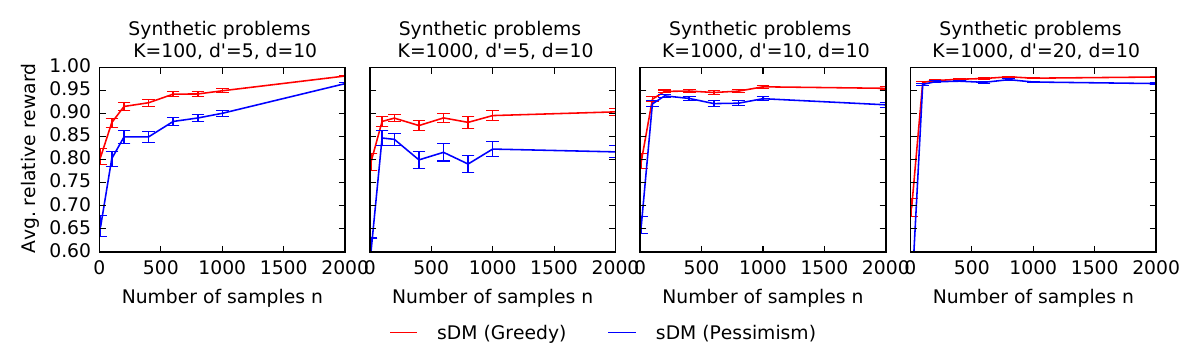}
  \caption{Comparison of \alg with greedy policy and \alg with pessimistic policy in OPL: The average MSE of an $\epsilon$-greedy policy on synthetic problems with varying $n$, $K$, and $d^\prime$.} 
  \label{fig:greedy_pessimism_results}
\end{figure}

\subsection{KuaiRec}\label{app:kuairec}

\textbf{Setting.} We use the KuaiRec dataset \citep{gao2022kuairec} to evaluate our method. This dataset contains two matrices:
\begin{enumerate}
    \item \texttt{big\_matrix}: A sparse matrix containing user-item interactions (watch ratios for various videos).
    \item \texttt{small\_matrix}: A smaller, fully observed dataset with complete user-item interactions. This matrix is only used for data collection and algorithm evaluation.
\end{enumerate}
The \texttt{big\_matrix} is used to obtain low-rank factorization embeddings through Singular Value Decomposition (SVD), representing users (contexts) and videos (actions). The \texttt{small\_matrix} is used to collect the sample set and to evaluate the algorithms. 

Before applying SVD, we preprocess \texttt{big\_matrix} as follows. First, we clipped the watch ratio scores to a maximum value of 10 to mitigate the impact of outliers and extreme values. Second, we normalized the interaction scores per user by dividing each user's interaction values by their maximum watch ratio. This scaling ensured that all user interaction scores ranged between 0 and 1, providing a uniform scale across different users.

After preprocessing, SVD was applied to the normalized matrix, resulting in video embeddings and user embeddings. The user embeddings are provided as input to the algorithms (serving as contexts), while the video embeddings are used as prior information by Bayesian algorithms that incorporate such priors. The dataset also includes category information for each video, which is beneficial for methods that rely on item categories.

\begin{figure}[H]
  \centering  \includegraphics[width=\linewidth]{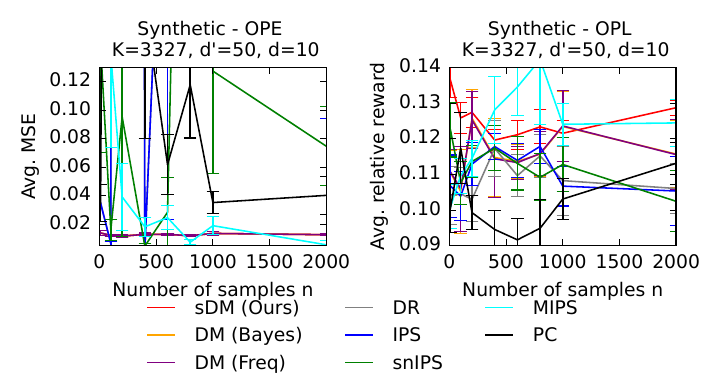}
  \caption{Comparison of \alg with baselines on the KuaiRec dataset for both OPE and OPL tasks.} 
  \label{fig:kauirec_ope_opl}
\end{figure}

\textbf{Results.} Since the rewards are collected from the \texttt{small\_matrix}, the true reward distribution is unknown, and there is no explicit prior from which videos are sampled. This characteristic makes the dataset a challenging benchmark for Bayesian algorithms like \alg, as it does not conform to the typical assumptions made by these methods. Despite this challenge, the strong performance of \alg in \cref{fig:kauirec_ope_opl} is particularly noteworthy. Specifically, \alg and \texttt{MIPS} demonstrate comparable and superior performance to other baselines in OPL, while \alg outperforms the baselines in OPE. It is important to note that results for OPE are hard to read since the performance of IPS-based methods fluctuated significantly when the number of samples $n$ changes  in this scenario.

\section{BROADER IMPACT}\label{app:impact}
This work contributes to the development and analysis of practical algorithms for offline learning to act under uncertainty. Our generic setting and algorithms have broad potential use, practitioners will therefore need to specifically address possible social impacts with respect to the relevant application.

\section{AMOUNT OF COMPUTATION REQUIRED}\label{app:compute}

Our experiments were conducted on internal machines with 30 CPUs and 1 GPU. Therefore, they required a moderate amount of computation.

\end{document}